\documentclass{article}

\usepackage{arxiv}
\usepackage[T1]{fontenc}
\usepackage{lmodern}
\usepackage{amsmath,amssymb}
\usepackage{graphicx}
\usepackage{booktabs}
\usepackage{array}
\usepackage{float}
\usepackage{microtype}
\usepackage{url}
\usepackage[hidelinks]{hyperref}
\usepackage{natbib}
\usepackage{authblk}
\setcitestyle{authoryear,round}
\pdftrailerid{}
\hypersetup{pdfcreator={},pdfproducer={}}
\providecommand{\MagpiePrimaryCSIZeroOne}{0.371}
\providecommand{\MagpiePrimaryCSIOneTwo}{0.304}
\providecommand{\MagpiePrimaryCSITwoThree}{0.238}
\providecommand{\MagpieDetectedEpisodeCount}{34{,}053}
\providecommand{\MagpieEpisodeDetectionRate}{65.1}
\providecommand{\MagpieEpisodeMeanLeadTime}{64.6}
\providecommand{\NoCIDetectedEpisodeCount}{29{,}838}
\providecommand{\NoCIEpisodeDetectionRate}{57.1}
\providecommand{\NoCIEpisodeMeanLeadTime}{59.2}
\providecommand{\NowcastNetEpisodeDetectionRate}{23.6}
\providecommand{\NowcastNetEpisodeMeanLeadTime}{18.3}
\providecommand{\LowAntecedentEpisodeCount}{9{,}843}
\providecommand{\MagpieLowAntecedentDetectionRate}{51.9}
\providecommand{\MagpieLowAntecedentMeanLeadTime}{38.5}
\providecommand{\NowcastNetLowAntecedentDetectionRate}{18.6}
\providecommand{\NowcastNetLowAntecedentMeanLeadTime}{11.5}
\providecommand{\GridBackboneOverallMeanCSI}{0.0154}
\providecommand{\BilinearEventOverallMeanCSI}{0.1331}
\providecommand{\MagpieOverallMeanCSI}{0.1433}
\providecommand{\GASetConvRelativeGain}{7.7}

\newcolumntype{P}[1]{>{\raggedright\arraybackslash}p{#1}}

\title{MAGPIE-Net: Predicting short-duration heavy-rainfall events in station neighborhoods from multitemporal FY-4A AGRI observations}

\author[1]{Xiang Lin}
\author[1,2]{Yunying Li\thanks{Corresponding author: \texttt{ghlyy@mail.iap.ac.cn}}}
\author[3,4]{Chengzhi Ye}
\author[1]{Zitong Chen}
\author[1]{Jing Sun}

\affil[1]{College of Meteorology and Oceanography, National University of Defense Technology, Changsha 410073, China}
\affil[2]{Key Laboratory of High Impact Weather (Special), China Meteorological Administration, Changsha 410073, China}
\affil[3]{Hunan Key Laboratory of Meteorological Disaster Prevention and Reduction, Changsha 410118, China}
\affil[4]{Institute of Meteorological Science of Hunan Province, Changsha 410118, China}

\date{}
\renewcommand{\headeright}{Preprint}
\renewcommand{\undertitle}{Preprint}
\renewcommand{\shorttitle}{MAGPIE-Net}

\begin{document}

\maketitle

\begin{abstract}
Short-duration heavy-rainfall warning determines whether 1 h rainfall will exceed a threshold within a target-station neighborhood over the next few hours. Multitemporal infrared and water-vapor observations from the Fengyun-4A Advanced Geostationary Radiation Imager (FY-4A AGRI) capture cloud-top cooling, moisture evolution, and cloud expansion before substantial surface rainfall develops. However, most deep-learning nowcasting methods convert these signals into local warnings by post-processing gridded precipitation predictions, preventing station-neighborhood event targets from directly supervising the satellite-to-station learning pathway. We propose MAGPIE-Net, which embeds a geographically adaptive, differentiable grid-to-station mapping in a pathway combining convection-initiation features, multiscale encoding, and auxiliary gridded precipitation diagnosis. Station-neighborhood event losses thereby constrain the satellite representation and its mapping to irregular station locations for 0--3 h event prediction. In independent 2023 warm-season tests over central and eastern China, critical success index (CSI) values under the primary $40\,\mathrm{km}/20\,\mathrm{mm\,h^{-1}}$ definition were \MagpiePrimaryCSIZeroOne{}, \MagpiePrimaryCSIOneTwo{}, and \MagpiePrimaryCSITwoThree{} at 0--1, 1--2, and 2--3 h. Across episodes, MAGPIE-Net achieved a detection rate of \MagpieEpisodeDetectionRate{}\% and a mean lead time of \MagpieEpisodeMeanLeadTime{} min, compared with \NowcastNetEpisodeDetectionRate{}\% and \NowcastNetEpisodeMeanLeadTime{} min for the best gridded-output baseline, and remained superior for smaller neighborhoods and the $50\,\mathrm{mm\,h^{-1}}$ threshold. During the critical early-warning stage, when antecedent 1 h rainfall within 40 km remained below 1 mm, MAGPIE-Net detected \MagpieLowAntecedentDetectionRate{}\% of episodes with a mean lead time of \MagpieLowAntecedentMeanLeadTime{} min. These results show that event-oriented satellite-to-station modeling converts multitemporal geostationary cloud and moisture observations into local heavy-rainfall warnings more effectively than gridded-precipitation modeling.
\end{abstract}


\section{Introduction}
\label{sec:introduction}

Short-duration heavy rainfall develops rapidly and can trigger damaging flash floods and urban flooding, particularly in small catchments and densely populated areas~\citep{fowler2021shortduration,westra2014short}. Over eastern China, extreme hourly rainfall is highly localized, and warm-season events are often associated with mesoscale convective systems~\citep{zhang2011hourly,chen2013sdhr}. Urban drainage systems and small catchments can respond within hours, making hourly rainfall monitoring and warning central to severe-convective-weather operations~\citep{zheng2015severe}. This task may become increasingly important as hourly precipitation extremes intensify in a warming climate~\citep{prein2017hourly}. Operational precipitation nowcasting has long relied on weather radar, whose frequent, high-resolution scans resolve the location, intensity, and movement of precipitation systems~\citep{wilson1998nowcasting,germann2002predictability,imhoff2022}. During the earliest stages of convective development, however, radar echoes may remain weak after cumulus clouds begin to grow vertically. A distinct precipitation echo may emerge only after rapid cloud development has begun~\citep{roberts2003storm}.

Geostationary satellite imagers can observe cloud development before precipitation echoes become pronounced. Infrared and water-vapor image sequences capture cloud-top cooling, vertical growth, horizontal expansion, glaciation-related spectral changes, and moistening in the middle and upper troposphere. These signals provide radiative indicators of convective development and have supported convection-initiation (CI) nowcasting across successive generations of geostationary satellites~\citep{mecikalski2006ci,sieglaff2011ci,walker2012ci,lee2017ci}. Early studies linked cloud-top cooling and multispectral changes to first radar echoes. The associated cloud-growth signals could precede the first significant radar echo by approximately 30--45 min~\citep{roberts2003storm,mecikalski2010infrared}. Later systems incorporated object tracking and machine-learning detection~\citep{zinner2008cbtram,han2019ci}. Recent Himawari-8 studies produced 30-min CI forecasts and detected local convective storms approximately 30 min to 1 h before radar in representative South China cases~\citep{li2024ciunet,yang2024ciSouthChina}. These studies establish satellite-observed cloud evolution as an observational basis for earlier warning while convection is still developing.

Geostationary satellites complement this early observational capability with broader spatial coverage. Effective weather-radar coverage can be constrained by network distribution, terrain blockage, and increasing beam height with range, particularly over offshore waters, mountainous terrain, and remote regions~\citep{saltikoff2019,lee2021goesConvection}. In contrast, geostationary infrared and water-vapor channels continuously observe broad land and ocean domains during both day and night. The Fengyun-4A Advanced Geostationary Radiation Imager (FY-4A AGRI) provides frequent multispectral information on cloud-top temperature, atmospheric moisture, and cloud-phase-related structure~\citep{yang2017fy4}. These characteristics allow AGRI to provide continuous information on convective development where radar coverage is limited and make it suitable for examining how regional cloud and moisture evolution supports local warning.

Broad observational availability, however, does not remove the physical difficulty of converting satellite-observed cloud signals into surface-rainfall warnings. Reviews of satellite precipitation retrieval identify the indirect radiance--rainfall relationship as a fundamental source of uncertainty~\citep{stephens2007remote,kidd2011status}. Surface rainfall development depends on updraft strength, cloud microphysics, low-level moisture supply, boundary-layer convergence, vertical wind shear, terrain forcing, and storm organization~\citep{doswell1996flash,houze2012orographic,houze2014cloud}. Similar cloud-top cooling can therefore accompany different rainfall outcomes. Storm motion and vertical wind shear can also displace the surface rainfall maximum from the coldest cloud top. Analyses over China show substantial differences between the spatial extent of convective cold clouds and the associated surface rainfall~\citep{chen2025coldcloud}. Satellite observations indicate convective development without uniquely specifying the location, intensity, or spatial structure of subsequent surface rainfall. A satellite-based warning model must therefore identify the cloud and moisture evolution patterns associated with subsequent heavy rainfall while allowing the warning target to accommodate spatial displacement between cloud structures and surface rain cores.

Based on this consideration, we formulate the warning target as a station-neighborhood heavy-rainfall event. An event occurs when 1 h accumulated rainfall reaches or exceeds a prescribed threshold within a specified radius of a target station. A single rain gauge samples only a small part of a convective system, while an intense rain core can shift as the system develops and moves. Future rainfall fields may differ in the exact location and fine-scale morphology of their rain cores. They nevertheless represent the same station-neighborhood event when threshold-exceeding rainfall occurs within the same target-station neighborhood. The event target therefore focuses prediction on whether the area surrounding a station will be affected while accommodating uncertainty in rain-core location.

This definition accords with neighborhood, scale-aware, and object-based verification methods for high-resolution precipitation forecasts~\citep{ebert2008fuzzy,roberts2008scale,gilleland2009spatial,davis2006object,gilleland2010verifying}. These methods accommodate small spatial displacements instead of treating them as complete forecast failures. Point-to-area sensitivity studies have similarly examined how neighborhood size affects short-duration heavy-rainfall verification~\citep{tian2015pointarea}. Operational practice provides direct precedent. China's national standard GB/T 44213-2024 recommends circular-neighborhood fuzzy verification and identifies 40 km as the preferred radius~\citep{sac2024severe}. Earlier descriptions of National Meteorological Center operations document the same station-centered 40-km point-to-area method~\citep{zheng2015severe}. The U.S. Weather Prediction Center likewise expresses excessive-rainfall risk as the probability of exceeding flash-flood guidance within 40 km of a point~\citep{erickson2021ero,burke2023ero}. These approaches preserve a location-specific warning target while accommodating displacement and scale uncertainty in localized convective rainfall.

Despite the operational relevance of station-neighborhood events, the prevailing deep-learning formulation for precipitation nowcasting remains gridded. Early radar-nowcasting systems framed prediction as convolutional sequence modeling or image-to-image translation~\citep{shi2015convlstm,shi2017trajgru,ayzel2020rainnet,trebing2021smaat}. Subsequent probabilistic and longer-range systems continued to generate spatial precipitation fields~\citep{sonderby2020metnet,pulkkinen2019pysteps,ravuri2021dgmr,espeholt2022metnet2}. Recent Transformer-based, physics-conditioned, and satellite-driven methods retain the same gridded formulation~\citep{earthformer2022,nowcastnet2023,park2025npm}. These models describe the spatial distribution and evolution of precipitation, but their prediction endpoint couples several requirements. The model must determine whether heavy rainfall develops while resolving its location, intensity, morphology, and movement. Satellite cloud-top and water-vapor observations constrain these properties unevenly. For localized and relatively rare heavy-rainfall events, neighborhood threshold exceedance is only one property that the predicted field must reproduce. Station-neighborhood warnings are then derived after training through thresholding, interpolation, or neighborhood aggregation. Training consequently optimizes the gridded precipitation field, whereas the final warning decision concerns a station-neighborhood event. With grid-to-station conversion outside the trained model, event-prediction errors cannot directly constrain satellite feature learning or the spatial mapping to individual stations.

\begin{figure}[htbp]
\centering
\includegraphics[width=\linewidth]{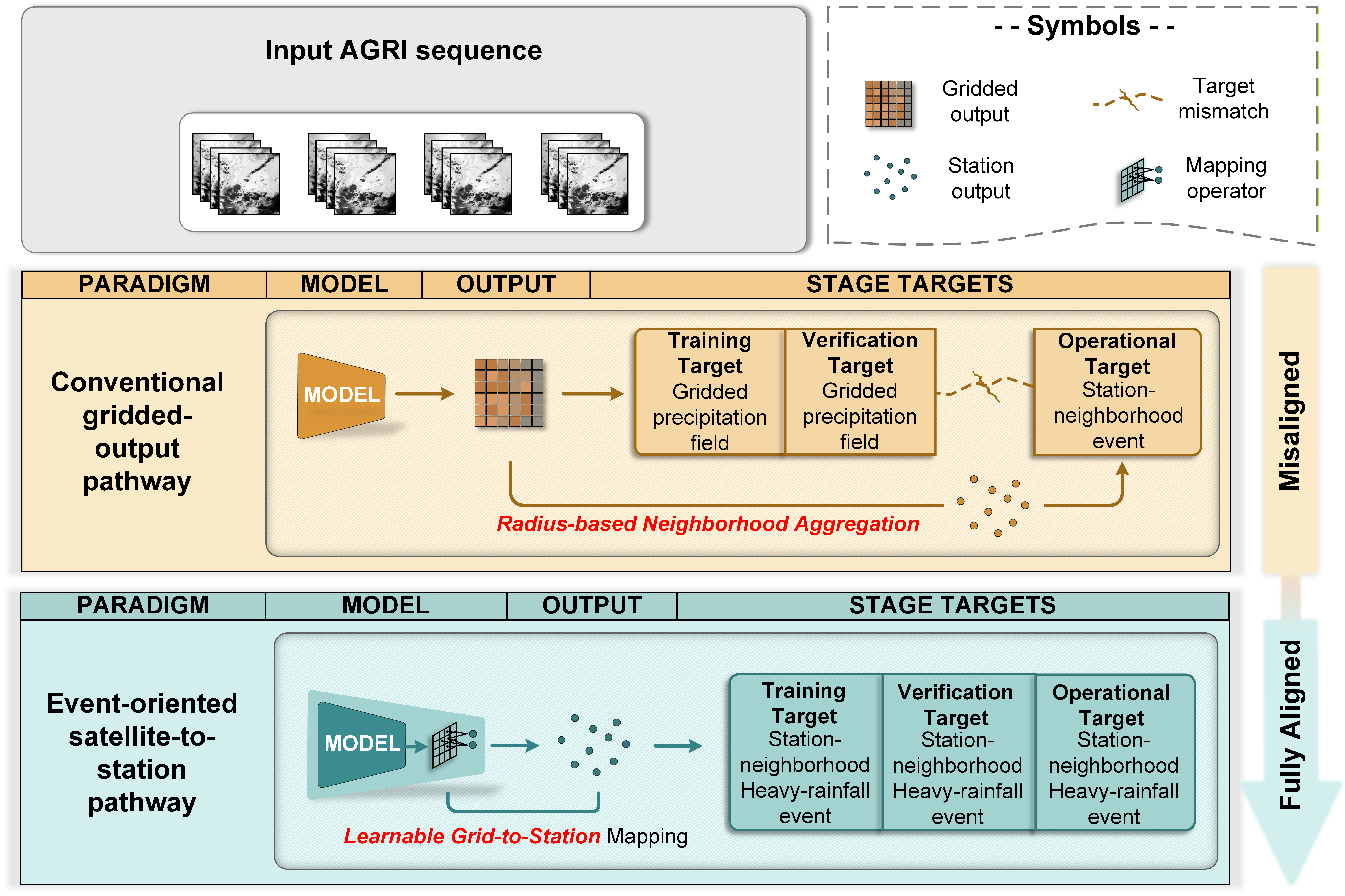}
\caption{Comparison of the conventional gridded-output pathway and the event-oriented satellite-to-station modeling pathway for satellite-based short-duration heavy-rainfall warning. Both pathways use regional multichannel geostationary satellite sequences. In the conventional pathway, model training and verification are based on gridded precipitation, and station-neighborhood events are subsequently derived through radius-based neighborhood aggregation. The event-oriented pathway incorporates a learnable grid-to-station mapping and uses station-neighborhood heavy-rainfall events as the common target for training, verification, and warning output.}
\label{fig:paradigm_comparison}
\end{figure}

Figure~\ref{fig:paradigm_comparison} contrasts the two pathways. The conventional approach derives station-neighborhood events from a gridded precipitation forecast after model training. The event-oriented pathway instead incorporates a learnable grid-to-station mapping and station-neighborhood event supervision into the model, allowing the final event loss to shape both the regional satellite representation and its mapping to target stations.

To implement this formulation, we develop MAGPIE-Net for predicting station-neighborhood heavy-rainfall events over the next 0--3 h from multitemporal FY-4A AGRI observations. The model derives CI features from rapidly developing cloud regions and encodes regional cloud and moisture evolution at multiple scales. An auxiliary diagnosis head projects the regional satellite representation onto three forecast-window-specific gridded precipitation fields. The geographically adaptive SetConv (GA-SetConv) decoder maps these fields to station-local representations using location- and lead-dependent spatial support while incorporating geographic attributes and cyclical time encodings. A multi-head module converts the resulting representations into event probabilities for multiple neighborhood radii and rainfall thresholds. Joint training propagates station-neighborhood event losses through the grid-to-station mapping, directly linking satellite representation learning to the final warning target.

We compare MAGPIE-Net with gridded-output baselines on an independent 2023 warm-season test set over central and eastern China, using matched station-neighborhood event definitions. The analysis spans forecast windows, event definitions, and stages of rainfall development, while controlled configurations identify component contributions. These comparisons assess whether the event-oriented satellite-to-station pathway translates cloud and moisture evolution in multitemporal AGRI observations into station-neighborhood heavy-rainfall event probabilities more effectively than the conventional gridded-output pathway.

\section{Data}
\label{sec:data_task}

\subsection{Study Region}
\label{subsec:study_area}

\begin{figure}[htbp]
\centering
\includegraphics[width=\linewidth]{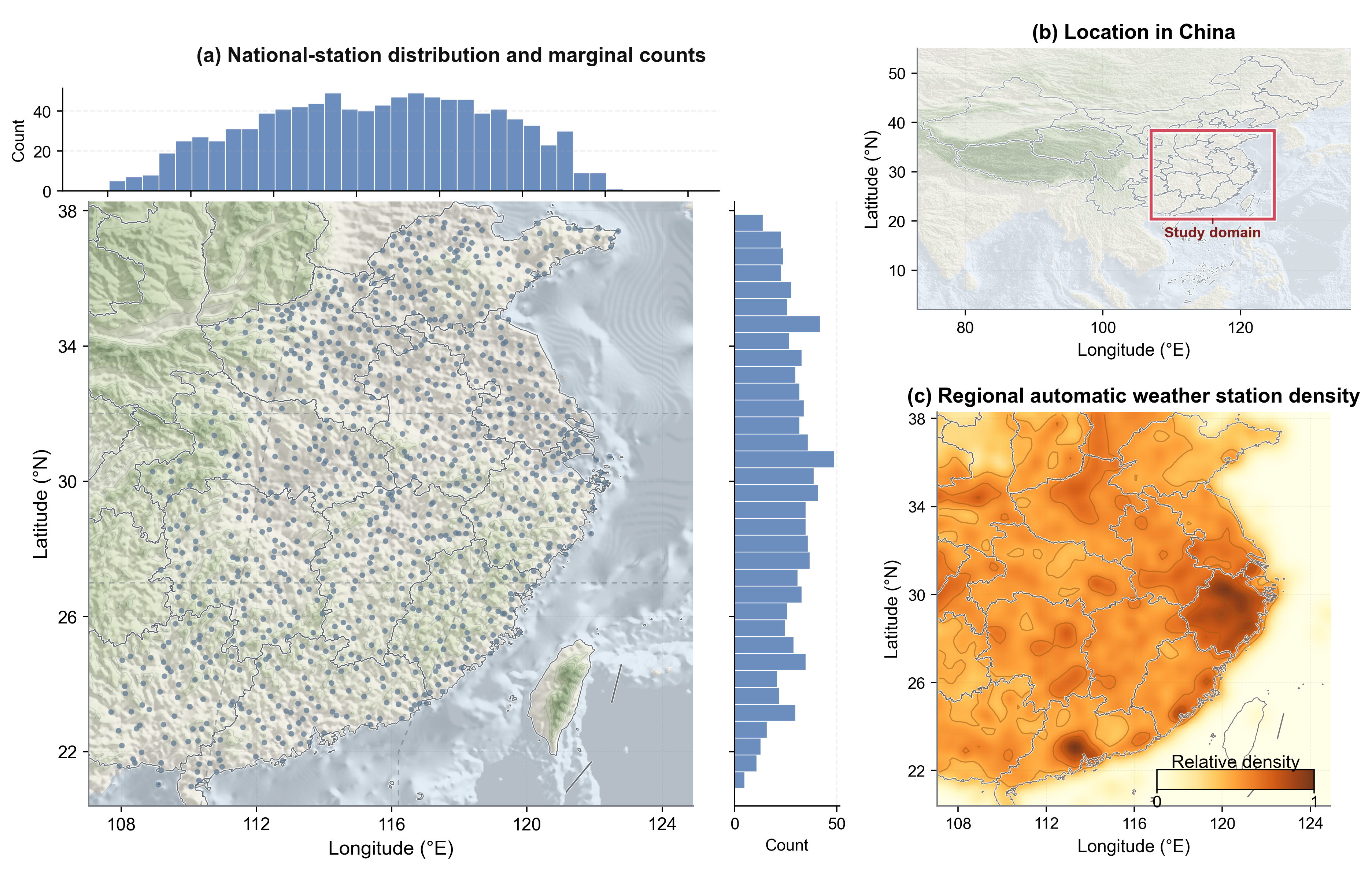}
\caption{Study domain and station networks. (a) Distribution of national meteorological stations, with marginal counts by longitude and latitude. (b) Location of the study domain within China. (c) Gaussian-smoothed relative density of regional automatic weather stations in the precipitation dataset.}
\label{fig:study_area_stations}
\end{figure}

The study domain spans central and eastern China from $20.40^\circ$ to $38.28^\circ$N and from $107.04^\circ$ to $124.92^\circ$E (Fig.~\ref{fig:study_area_stations}). It encompasses inland and coastal regions, plains, and mountainous terrain. Warm-season short-duration heavy rainfall is associated with frontal and warm-sector convection, mesoscale convective systems, tropical-cyclone rainbands, and terrain-related forcing~\citep{chen2013sdhr,luo2017scmrex,lonfat2004tc,houze2012orographic}. This combination of precipitation regimes and terrain settings provides a suitable setting for evaluating satellite-based station-neighborhood heavy-rainfall warning.

\subsection{FY-4A AGRI Observations}
\label{subsec:fy4a_agri}

The satellite input is derived from FY-4A AGRI products provided by the National Satellite Meteorological Center of the China Meteorological Administration. We use channels 7--14, which span the mid-wave infrared, water-vapor, thermal-infrared window, split-window, and CO$_2$ absorption bands and are available during both day and night. Together, these channels describe cloud-top temperature, cloud phase, and moisture in the middle and upper troposphere. Changes between successive scans capture cloud-top cooling, cloud growth, horizontal expansion, and moisture evolution, thereby describing the evolution of regional cloud and moisture fields~\citep{yang2017fy4,mecikalski2010infrared}. Because the radiance response of each channel also depends on cloud vertical structure and optical and microphysical properties~\citep{sun2025agriCloud}, the eight channels are used jointly as multitemporal predictors. No individual channel is treated as a direct indicator of surface rainfall.

Table~\ref{tab:agri_channels} summarizes the spectral channels used as model input. After geolocation and spatial registration, all observations are remapped to a common $448\times448$ latitude--longitude grid covering the study domain, with an approximate spacing of $0.04^\circ$. Each AGRI scan is therefore represented by eight spatially aligned channel fields on the regional grid.

\begin{table}[htbp]
\centering
\caption{FY-4A AGRI channels used as model input.}
\label{tab:agri_channels}
\footnotesize
\setlength{\tabcolsep}{4pt}
\begin{tabular}{
>{\centering\arraybackslash}p{0.11\linewidth}
>{\centering\arraybackslash}p{0.35\linewidth}
>{\centering\arraybackslash}p{0.20\linewidth}
>{\centering\arraybackslash}p{0.18\linewidth}}
\toprule
Channel & Type & Central wavelength & Native spatial resolution \\
\midrule
7  & Mid-wave infrared (high gain) & $3.75\,\mu\mathrm{m}$ & 2 km\textsuperscript{a} \\
8  & Mid-wave infrared (low gain)  & $3.75\,\mu\mathrm{m}$ & 4 km \\
9  & Upper-tropospheric water vapor & $6.25\,\mu\mathrm{m}$ & 4 km \\
10 & Mid-tropospheric water vapor  & $7.10\,\mu\mathrm{m}$ & 4 km \\
11 & Long-wave infrared            & $8.50\,\mu\mathrm{m}$ & 4 km \\
12 & Thermal-infrared window       & $10.70\,\mu\mathrm{m}$ & 4 km \\
13 & Split-window infrared         & $12.00\,\mu\mathrm{m}$ & 4 km \\
14 & CO$_2$ absorption             & $13.50\,\mu\mathrm{m}$ & 4 km \\
\bottomrule
\end{tabular}
\vspace{1mm}

\begin{minipage}{0.94\linewidth}
\raggedright
\footnotesize\textsuperscript{a}\,Channel 7 has a native spatial resolution of 2 km, but the official AGRI product used here provides this channel at 4 km. All selected channels are subsequently remapped to the common grid with an approximate spacing of $0.04^\circ$.
\end{minipage}
\end{table}

\subsection{Ground-Station Precipitation Observations}
\label{subsec:gauge_observations}

Ground-station precipitation observations were obtained from the National Meteorological Information Center of the China Meteorological Administration. The dataset reports 1 h accumulated precipitation at 15 min intervals for national meteorological stations and regional automatic weather stations. At each valid time $t$, the available record includes every station with a valid observation, its precipitation accumulation over the preceding hour, and its longitude, latitude, and elevation. We denote the set of valid stations by $\mathcal{S}_t$. For station $s$, $P_s(t)$ denotes the 1 h accumulated precipitation, and $r_s=(\lambda_s,\phi_s,h_s)$ contains the geographic attributes. These records form the basis of the station-neighborhood heavy-rainfall event labels.

\subsection{Station-Neighborhood Heavy-Rainfall Event Definition}
\label{subsec:station_neighborhood_label}

For a target station $s$, consider the 1 h window ending at time $t$. A station-neighborhood heavy-rainfall event occurs when at least one station within radius $R$ of $s$ records a 1 h accumulation that meets threshold $q$. The neighborhood calculation includes every national meteorological station and regional automatic weather station with a valid rainfall record at $t$. For $s\in\mathcal{S}_t$, the neighborhood is defined as
\begin{equation}
\mathcal{N}_R(s,t)
=
\left\{
 s_i\in\mathcal{S}_t
 \mid d(s_i,s)\le R
\right\}
\label{eq:station_neighborhood}
\end{equation}
where $d(s_i,s)$ denotes the geographic distance between stations $s_i$ and $s$. The maximum 1 h accumulated precipitation within the neighborhood is
\begin{equation}
P_s^{R,\max}(t)
=
\max_{s_i\in\mathcal{N}_R(s,t)} P_{s_i}(t)
\label{eq:station_neighborhood_max}
\end{equation}
and the corresponding binary event label is
\begin{equation}
 y_s^{R,q}(t)
=
\begin{cases}
1, & P_s^{R,\max}(t)\ge 20,\quad q=20,\\
1, & P_s^{R,\max}(t)>50,\quad q=50,\\
0, & \mathrm{otherwise}.
\end{cases}
\label{eq:station_event_label}
\end{equation}
Thus, $y_s^{R,q}(t)=1$ indicates that the event criterion is satisfied at target station $s$ during the 1 h window ending at $t$.

Three neighborhood radii, $R\in\{40,20,10\}\,\mathrm{km}$, are paired with two rainfall thresholds, $q\in\{20,50\}\,\mathrm{mm\,h^{-1}}$, to produce six event definitions. The complete label map at time $t$ is written as
\begin{equation}
\mathbf{Y}(t)
=
\left\{
\left(
 r_s,
 \left\{y_s^{R,q}(t)\right\}_{R\in\{40,20,10\},\,q\in\{20,50\}}
\right)
\mid s\in\mathcal{S}_t
\right\}
\label{eq:station_label_map}
\end{equation}
At each valid time, $\mathbf{Y}(t)$ forms an irregular station-based label map containing the geographic attributes and six binary event labels for every valid station. The $40\,\mathrm{km}/20\,\mathrm{mm\,h^{-1}}$ combination is used as the primary event definition. The remaining five definitions serve as auxiliary targets during joint training and support evaluation under stricter spatial or rainfall-intensity criteria. Smaller radii require more precise event localization, whereas the $50\,\mathrm{mm\,h^{-1}}$ threshold focuses on more intense short-duration rainfall.

\subsection{Sample Construction and Data Split}
\label{subsec:sample_construction_split}
\label{subsec:sample_construction_imbalance}

Each forecasting sample is defined by an initialization time $t_0$. The satellite input consists of four AGRI scans acquired at 15 min intervals before initialization:
\begin{equation}
\mathcal{T}_{t_0}^{\mathrm{sat}}
=
\left\{
 t_0-60\,\mathrm{min},
 t_0-45\,\mathrm{min},
 t_0-30\,\mathrm{min},
 t_0-15\,\mathrm{min}
\right\}
\label{eq:sat_time_set_data}
\end{equation}
Let $\mathbf{B}(t)\in\mathbb{R}^{8\times H\times W}$ denote the eight-channel AGRI field at time $t$. The four scans are concatenated along the channel dimension to form the satellite input:
\begin{equation}
X_{t_0}^{\mathrm{sat}}
=
\operatorname{Concat}_{\mathrm{ch}}
\left[
\mathbf{B}(t)
\right]_{t\in\mathcal{T}_{t_0}^{\mathrm{sat}}}
\in
\mathbb{R}^{32\times H\times W}
\label{eq:satellite_input_sequence}
\end{equation}
The three forecast windows are
\begin{equation}
H_k
=
\left(t_0+(k-1)\,\mathrm{h},\,t_0+k\,\mathrm{h}\right]
\qquad k=1,2,3
\label{eq:forecast_windows}
\end{equation}
and the corresponding targets are the station-neighborhood event-label maps ending 1, 2, and 3 h after initialization:
\begin{equation}
\mathbf{Y}_{t_0}^{\mathrm{tar}}
=
\left[
\mathbf{Y}(t_0+1\,\mathrm{h}),
\mathbf{Y}(t_0+2\,\mathrm{h}),
\mathbf{Y}(t_0+3\,\mathrm{h})
\right]
\label{eq:sample_target_maps}
\end{equation}
Since each label map is constructed from the 1 h accumulated precipitation ending at its valid time, the three targets correspond to the 0--1, 1--2, and 2--3 h forecast windows, respectively.

Station masks distinguish the locations used as prediction targets from those used only as neighborhood observations. National meteorological stations have more consistent quality control and temporal coverage, so they are used for supervision during training and validation and for the primary independent-test evaluation. Regional automatic weather stations contribute to the neighborhood rainfall observations used to construct the event labels but are excluded from supervised targets and primary national-station metrics. The modeling domain contains 987 national stations, of which 973 within the central evaluation crop are retained for the primary test metrics to reduce boundary effects.

Samples are restricted to May through September and split strictly by calendar year. Data from 2018--2021 are used for training, data from 2022 for validation and hyperparameter selection, and data from 2023 for independent testing. This temporal separation keeps the entire 2023 warm season outside model fitting and selection. To reduce the dominance of event-free conditions during training, we retain 10\% of samples with no positive event label across the supervised target maps and all samples containing at least one positive label. The validation and test sets retain their original distributions. The final training, validation, and independent test sets contain 26,121, 3,136, and 3,097 samples, respectively.

\section{MAGPIE-Net: An Event-Oriented Satellite-to-Station Architecture}
\label{sec:method}

\subsection{Overall Architecture}
\label{subsec:overall_framework}

\begin{figure}[htbp]
\centering
\includegraphics[width=\linewidth]{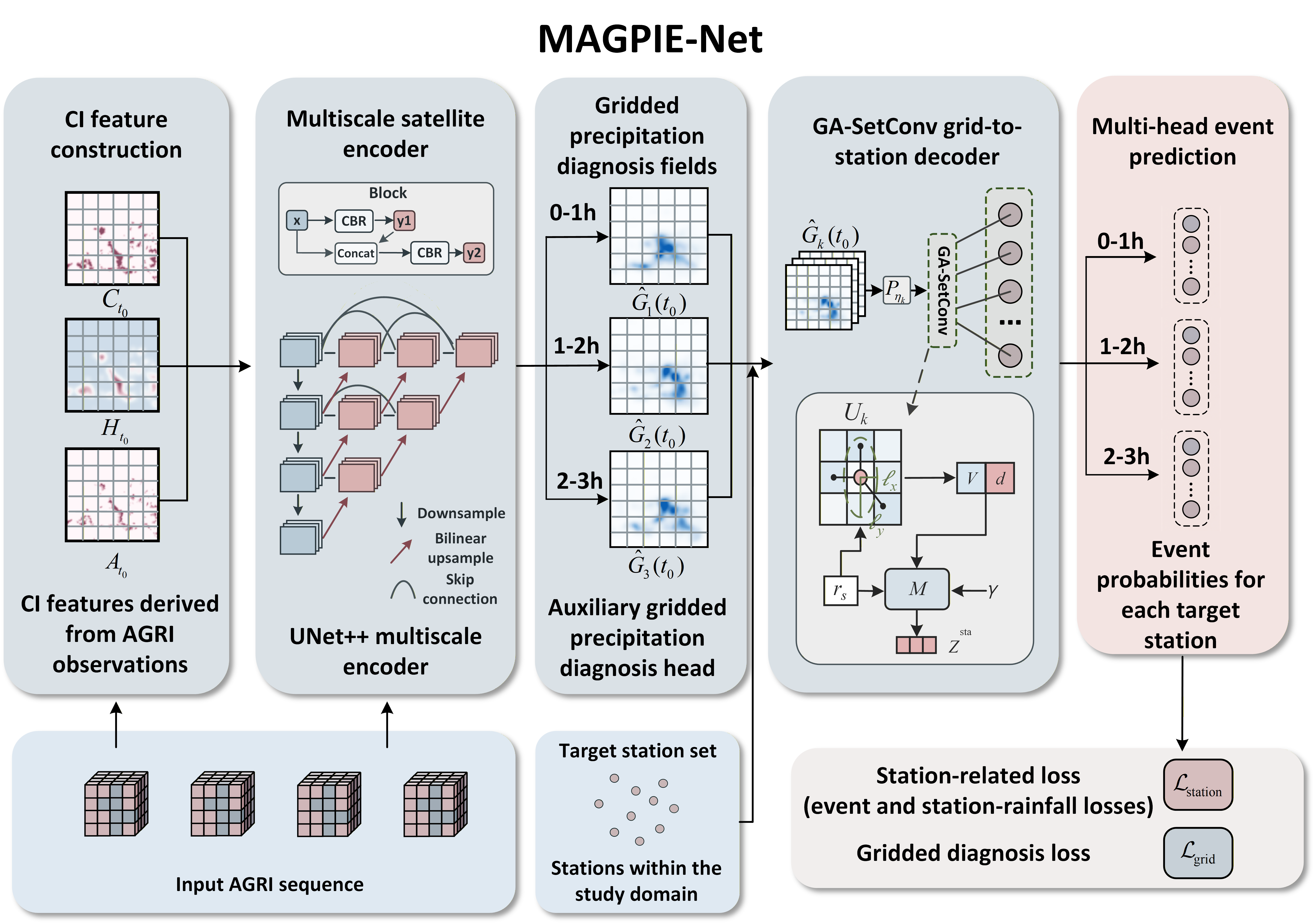}
\caption{Overall architecture of MAGPIE-Net. The input AGRI sequence is used to construct CI features, which are combined with the original AGRI channels by the multiscale satellite encoder. The encoder produces a regional satellite representation, from which the auxiliary diagnosis head generates three gridded precipitation fields. GA-SetConv maps these fields to target stations, and the multi-head module predicts event probabilities for three forecast windows. CBR denotes a convolution--batch-normalization--ReLU block. For clarity, repeated indices are omitted from the GA-SetConv inset; the complete notation is defined in Section~\ref{subsec:ga_setconv_decoder}.}
\label{fig:magpie_framework}
\end{figure}

Figure~\ref{fig:magpie_framework} summarizes the overall architecture of MAGPIE-Net. The model comprises a convection-initiation (CI) feature-construction module, a multiscale satellite encoder, an auxiliary gridded precipitation diagnosis head, a GA-SetConv grid-to-station decoder, and a multi-head event prediction module. It preserves cloud- and rainfall-related spatial structures on the regional grid before mapping them to irregular target stations. Because this mapping is differentiable, station-neighborhood event losses constrain both the shared satellite representation and its mapping to target stations.

For an initialization time $t_0$, the model receives the input AGRI sequence $X_{t_0}^{\mathrm{sat}}$ defined in Section~\ref{subsec:sample_construction_split} and a set of target stations $\mathcal{S}^{\mathrm{tar}}$. Let $\mathcal{R}=\{40,20,10\}\,\mathrm{km}$ and $\mathcal{Q}=\{20,50\}\,\mathrm{mm\,h^{-1}}$ denote the neighborhood radii and rainfall thresholds. The complete model mapping is expressed as
\begin{equation}
\begin{aligned}
\left[
\widehat{\mathbf{P}}_{1}(t_0),
\widehat{\mathbf{P}}_{2}(t_0),
\widehat{\mathbf{P}}_{3}(t_0)
\right]
&=
\operatorname{MAGPIE}_{\Theta}
\left(
X_{t_0}^{\mathrm{sat}},
\mathcal{S}^{\mathrm{tar}}
\right),\\
\widehat{\mathbf{P}}_{k}(t_0)
&=
\left\{
\hat{p}_{s,k}^{R,q}(t_0)
\;\middle|\;
 s\in\mathcal{S}^{\mathrm{tar}},
 R\in\mathcal{R},
 q\in\mathcal{Q}
\right\},
\qquad k=1,2,3.
\end{aligned}
\label{eq:magpie_complete_mapping}
\end{equation}
For each forecast window $H_k$, $\widehat{\mathbf{P}}_{k}(t_0)$ contains event probabilities for every target station and radius--threshold combination. The $40\,\mathrm{km}/20\,\mathrm{mm\,h^{-1}}$ head corresponds to the primary event definition, while the remaining five heads represent stricter spatial or rainfall-intensity criteria. Each $\hat{p}_{s,k}^{R,q}(t_0)$ lies in $[0,1]$. Binary event predictions are obtained by applying decision thresholds selected on the validation set.

The CI feature-construction module identifies rapid cloud development in the AGRI sequence and represents its spatial extent and timing. The multiscale satellite encoder combines the original AGRI channels with the CI features to derive a regional satellite representation. The auxiliary diagnosis head projects this representation onto three gridded precipitation fields. GA-SetConv maps the gridded features to station-local representations and incorporates geographic attributes and cyclical time encodings. The multi-head event prediction module converts these representations into event probabilities.

\subsection{Convection-Initiation Feature Construction}
\label{subsec:ci_prior}

Rapid cloud-top cooling, together with spectral changes in infrared and water-vapor channels, is commonly associated with developing convection~\citep{mecikalski2006ci,mecikalski2010infrared,walker2012ci}. MAGPIE-Net represents these signals using motion-compensated pixel-level convection-initiation (CI) masks that identify localized regions of rapid cloud development. The detection criteria are based on the physical indicators used by Peng et al.~\citep{peng2025newci}, with relaxed thresholds to retain a broader candidate set.

CI masks are constructed for the two candidate times $\tau\in\{t_0-30\,\mathrm{min},t_0-15\,\mathrm{min}\}$. For each $\tau$, the AGRI observations at $\tau-30$, $\tau-15$, and $\tau$ are used. Bidirectional motion fields are estimated from the $10.7\,\mu\mathrm{m}$ images using the Dense Inverse Search (DIS) optical-flow algorithm in OpenCV with the medium preset. These fields trace each pixel at $\tau$ through the two preceding scans, so cooling is calculated along the estimated cloud trajectory rather than between fixed grid cells. Candidate pixels must also satisfy forward--backward flow consistency and displacement limits before the spectral and cooling criteria in Table~\ref{tab:ci_rule} are applied.

\begin{table}[htbp]
\centering
\caption{Parameters and criteria for constructing CI candidate masks.}
\label{tab:ci_rule}
\small
\setlength{\tabcolsep}{4pt}
\begin{tabular}{p{0.28\linewidth} p{0.62\linewidth}}
\toprule
Item & Specification or condition \\
\midrule
Candidate times &
$\tau=t_0-30\,\mathrm{min}$ and $t_0-15\,\mathrm{min}$, each evaluated using the three scans at $\tau-30$, $\tau-15$, and $\tau$ \\

Motion consistency &
Backward trajectories remain within the grid, displacement in each 15-min step is at most 12 pixels, and forward--backward error is at most 1.5 pixels \\

Trajectory notation &
$\widetilde{T}_{10.7}$ denotes a temperature sampled along the backward-traced cloud trajectory \\

Cold cloud top &
$T_{10.7}(\tau,g)<273\,\mathrm{K}$ \\

Continuous cooling &
$\widetilde{T}_{10.7}(\tau-15,g)-\widetilde{T}_{10.7}(\tau-30,g)\leq-4\,\mathrm{K}$ and
$T_{10.7}(\tau,g)-\widetilde{T}_{10.7}(\tau-15,g)\leq-4\,\mathrm{K}$ \\

Water-vapor--infrared difference &
$T_{7.1}(\tau,g)-T_{10.7}(\tau,g)>-30\,\mathrm{K}$ \\

Split-window difference &
$T_{12.0}(\tau,g)-T_{10.7}(\tau,g)>-3\,\mathrm{K}$ \\

Three-channel combination &
$T_{8.5}(\tau,g)+T_{12.0}(\tau,g)-2T_{10.7}(\tau,g)>-5\,\mathrm{K}$ \\

Spatial coherence &
The connected candidate component contains at least 4 pixels and covers at least $64\,\mathrm{km^2}$ \\
\bottomrule
\end{tabular}
\end{table}

The relatively permissive thresholds favor candidate retention, so some identified cores may not subsequently produce heavy rainfall. The masks serve as auxiliary inputs rather than event labels. Evaluated together with the full multispectral context, they allow the encoder to distinguish candidates associated with subsequent station-neighborhood heavy rainfall from less informative cloud-development signals.

The candidate masks obtained from the input AGRI sequence are merged into a single field:
\begin{equation}
C_{t_0}(g)
=
\max_{\tau} I_{\tau}(g)
\label{eq:ci_max}
\end{equation}
where $I_{\tau}(g)$ denotes the motion-compensated pixel-level CI mask at candidate time $\tau$. The merged field $C_{t_0}$ marks locations where rapid cloud development is detected at either of the two candidate times.

Cloud-top features and the associated surface-rainfall area may separate as convective systems grow and move~\citep{chen2025coldcloud}. A Gaussian filter is therefore applied to extend the candidate signal into the surrounding cloud region. The smoothed signal is represented by a three-channel field aligned with the three forecast windows:
\begin{equation}
H_{t_0}
=
\operatorname{Rep}_{3}
\left\{
\operatorname{Norm}
\left[
\mathcal{K}_{\sigma}*C_{t_0}
\right]
\right\}
\in
\mathbb{R}^{3\times H\times W}
\label{eq:ci_halo}
\end{equation}
where $\operatorname{Norm}[\cdot]$ scales the smoothed field to $[0,1]$, and $\operatorname{Rep}_{3}$ repeats the same normalized field across three channels, one for each forecast window. A Gaussian kernel with a radius of 10 pixels and $\sigma=5$ pixels is used. These channels provide the shared encoder with a common representation of the spatial context surrounding CI candidates. Forecast-window-specific responses to this context are learned by the subsequent preprocessing and station-decoding branches.

A timing field $A_{t_0}$ records the most recent CI detection at each grid cell. Its value is $2/3$ for a most recent detection at $t_0-30\,\mathrm{min}$, 1 for a detection at $t_0-15\,\mathrm{min}$, and 0 where neither mask contains a candidate. The merged candidate field, the spatially enhanced field, and the timing field are concatenated to form the CI feature set:
\begin{equation}
Z_{t_0}^{\mathrm{CI}}
=
\operatorname{Concat}_{\mathrm{ch}}
\left(
C_{t_0},
H_{t_0},
A_{t_0}
\right)
\in
\mathbb{R}^{5\times H\times W}
\label{eq:ci_prior_final}
\end{equation}

\subsection{Multiscale Encoding with Auxiliary Precipitation Supervision}
\label{subsec:encoder_grid_diagnostic}

Convective cloud and precipitation systems span a wide range of spatial scales. Rapidly cooling cloud tops and newly formed convective cores are localized, whereas cloud clusters, anvils, and the surrounding moisture environment extend over much larger areas~\citep{mecikalski2006ci,houze2014cloud}. The encoder must therefore retain fine-scale cloud-development features while also representing the broader environment that affects storm growth and rainfall organization.

The input AGRI sequence and CI features are concatenated along the channel dimension:
\begin{equation}
\widetilde{X}_{t_0}
=
\operatorname{Concat}_{\mathrm{ch}}
\left(
X_{t_0}^{\mathrm{sat}},
Z_{t_0}^{\mathrm{CI}}
\right)
\in
\mathbb{R}^{37\times H\times W}
\label{eq:concat_input}
\end{equation}
The multiscale encoder follows a UNet++ architecture with nested skip connections~\citep{ronneberger2015unet,zhou2018unetpp}. It uses four resolution levels with 64, 128, 256, and 512 feature channels. The nested skip connections combine broad cloud and moisture patterns from deeper levels with higher-resolution cloud structures to produce the regional satellite representation
\begin{equation}
F_{t_0}
=
E_{\theta}
\left(
\widetilde{X}_{t_0}
\right)
\label{eq:encoder}
\end{equation}

The regional representation is projected onto gridded precipitation diagnosis fields for the three forecast windows through an auxiliary $1\times1$ regression head:
\begin{equation}
\widehat{\mathbf{G}}(t_0)
=
D_{\phi}
\left(
F_{t_0}
\right)
=
\left[
\hat{G}_{1}(t_0),
\hat{G}_{2}(t_0),
\hat{G}_{3}(t_0)
\right]
\in
\mathbb{R}^{3\times H\times W}
\label{eq:grid_reg_head}
\end{equation}
Each diagnosis field represents precipitation-related spatial structure for one forecast window. Together, the three fields provide the gridded input to the station decoder.

The auxiliary reference field $G_k^{*}(t_0)$ is generated from station precipitation observations through Barnes objective analysis~\citep{barnes1964objective} on the same $0.04^\circ$ grid as the satellite input. This spatially continuous target describes the location and broad organization of observed surface rainfall. Supervision against this target encourages the regional satellite representation to preserve precipitation-related spatial structure for subsequent station-neighborhood event prediction.

\subsection{GA-SetConv: Geographically Adaptive Grid-to-Station Decoding}
\label{subsec:ga_setconv_decoder}

The encoder produces features on a regular grid, whereas the warning targets are defined at irregular station locations. GA-SetConv connects these representations through SetConv-style kernel aggregation, following permutation-invariant set representations and conditional neural processes for mappings between irregular samples and continuous fields~\citep{zaheer2017deepsets,garnelo2018cnp,gordon2019convolutional}. Rainfall organization varies with geographic setting, while storm motion and vertical wind shear can displace surface rainfall from the corresponding cloud-top features~\citep{houze2012orographic,houze2014cloud,chen2025coldcloud}. The required spatial support can therefore differ among target stations and forecast windows. GA-SetConv uses an axis-aligned learnable kernel whose east--west and north--south length scales are conditioned on the target-station attributes and forecast window.

The gridded precipitation diagnosis fields are first processed by a separate U-Net branch for each forecast window:
\begin{equation}
U_k(t_0)
=
P_{\eta_k}
\left(
\widehat{\mathbf{G}}(t_0)
\right),
\qquad k=1,2,3
\label{eq:decoder_preproc}
\end{equation}
where $U_k(t_0)$ contains the gridded features used to predict events in forecast window $H_k$. The three branches generate forecast-window-specific features, allowing the spatial mapping to represent changes in existing rainfall structure, storm movement, spatial expansion, and new-cell development across forecast windows.

Let $(x_g,y_g)$ and $(x_s,y_s)$ denote the normalized coordinates of grid cell $g$ and target station $s$, respectively. For each forecast window, a geographic scale network uses the station attributes $r_s=(\lambda_s,\phi_s,h_s)$ defined in Section~\ref{subsec:gauge_observations} to estimate the east--west and north--south length scales:
\begin{equation}
\begin{bmatrix}
\ell_{x,s,k}\\
\ell_{y,s,k}
\end{bmatrix}
=
\exp
\left[
 h_{\psi_k}(r_s)
\right]
\label{eq:geo_adaptive_axial_scale}
\end{equation}
The corresponding kernel is
\begin{equation}
K_{s,k}(g)
=
\exp
\left[
-\frac{(x_g-x_s)^2}{2\ell_{x,s,k}^{2}}
-\frac{(y_g-y_s)^2}{2\ell_{y,s,k}^{2}}
\right]
\label{eq:axial_rbf_kernel}
\end{equation}
The two length scales independently control the east--west and north--south support. The aggregation region can therefore vary by target station and forecast window while remaining axis aligned.

For each target station and forecast window, the total kernel weight and normalized station-local feature are
\begin{equation}
\begin{aligned}
d_{s,k}
&=
\sum_{g\in\mathcal{G}} K_{s,k}(g),\\
V_{s,k}
&=
\frac{
\sum_{g\in\mathcal{G}}K_{s,k}(g)U_k(t_0,g)
}{
d_{s,k}+\epsilon
}.
\end{aligned}
\label{eq:setconv_mapping}
\end{equation}
Dividing by the total kernel weight keeps the magnitude of $V_{s,k}$ comparable as the learned aggregation region expands or contracts. The total weight $d_{s,k}$ is retained as an additional input, allowing the station decoder to account for differences in the effective spatial support.

The normalized station-local feature $V_{s,k}$ and total kernel weight $d_{s,k}$ are combined with the station attributes $r_s$ and the cyclical encoding $\gamma_{t_0,k}$ of the valid time:
\begin{equation}
Z_{s,k}^{\mathrm{sta}}
=
M_{\xi_k}
\left(
V_{s,k},
 d_{s,k},
 r_s,
 \gamma_{t_0,k}
\right)
\label{eq:station_trunk_feature}
\end{equation}
The station attributes represent persistent spatial differences through longitude, latitude, and elevation, whereas the cyclical encoding describes seasonal and diurnal timing~\citep{chen2013sdhr}. Together with $V_{s,k}$ and $d_{s,k}$, these variables form the station-local representation for each forecast window.

The station-local representation is passed to six event heads:
\begin{equation}
\begin{aligned}
\ell_{s,k}^{R,q}
&=
C_{R,q}
\left(
Z_{s,k}^{\mathrm{sta}}
\right),\\
\hat{p}_{s,k}^{R,q}(t_0)
&=
\sigma
\left(
\ell_{s,k}^{R,q}
\right),
\qquad
R\in\mathcal{R},\ q\in\mathcal{Q}.
\end{aligned}
\label{eq:multiscale_station_event_probabilities}
\end{equation}
The $40\,\mathrm{km}/20\,\mathrm{mm\,h^{-1}}$ head produces the primary station-neighborhood heavy-rainfall event probability. The other five heads correspond to smaller neighborhoods and the higher rainfall threshold, providing additional constraints on spatial localization and rainfall intensity. A parallel regression branch supplies continuous station-rainfall information during training.

GA-SetConv therefore converts the gridded precipitation features derived from the shared regional representation into station-local representations, which the multi-head event prediction module converts into event probabilities at irregular station locations. Its axis-aligned spatial support adapts independently in the east--west and north--south directions for each target station and forecast window. Each forecast-window preprocessing branch uses a two-level U-Net with a base width of 32. The geographic scale network has two hidden layers with 32 units each, and the station decoder produces a 64-dimensional representation before the event heads.

\subsection{Multi-Task Training Objective}
\label{subsec:loss_function}

The final warning target is defined by station-neighborhood event labels, making the event loss the principal station-level supervision. Because binary labels describe threshold exceedance but not rainfall intensity or the spatial organization of the precipitation system, the training objective also includes gridded precipitation-diagnosis and station-rainfall regression losses. These auxiliary terms constrain regional rainfall structure and local intensity, respectively.

For neighborhood radius $R$ and rainfall threshold $q$, the event loss is defined using positive-weighted binary cross-entropy. Following the notation in Section~\ref{subsec:station_neighborhood_label}, the observed label for forecast window $H_k$ is $y_s^{R,q}(t_0+k\,\mathrm{h})$:
\begin{equation}
\mathcal{L}_{R,q}
=
\frac{
\sum_{t_0,s,k}
M_{s,k}(t_0)
\,
\mathrm{BCE}_{\alpha}
\left(
\ell_{s,k}^{R,q}(t_0),
 y_s^{R,q}(t_0+k\,\mathrm{h})
\right)
}{
\sum_{t_0,s,k}M_{s,k}(t_0)+\epsilon
}
\label{eq:station_bce_multiscale}
\end{equation}
where $M_{s,k}(t_0)$ indicates whether the corresponding station label is available, and
\begin{equation}
\mathrm{BCE}_{\alpha}(\ell,y)
=
-\alpha y\log\sigma(\ell)
-(1-y)\log\left[1-\sigma(\ell)\right]
\label{eq:weighted_bce}
\end{equation}
The positive-event weight is set to $\alpha=2$. Together with the downsampling of event-free training samples described in Section~\ref{subsec:sample_construction_imbalance}, this weighting gives greater influence to the less frequent positive events during optimization.

The losses from the six event heads are averaged:
\begin{equation}
\overline{\mathcal{L}}_{\mathrm{event}}
=
\frac{1}{6}
\sum_{R\in\mathcal{R}}
\sum_{q\in\mathcal{Q}}
\mathcal{L}_{R,q}
\label{eq:mean_event_loss}
\end{equation}
Joint optimization of the six event heads provides the shared station representation with supervision across neighborhood scales and rainfall thresholds, allowing the same representation to support all six event definitions.

The gridded diagnosis loss is
\begin{equation}
\mathcal{L}_{\mathrm{grid}}
=
\frac{
\sum_{t_0,k,g}
M_{k}^{\mathrm{grid}}(t_0,g)
\left[
\hat{G}_{k}(t_0,g)-G_{k}^{*}(t_0,g)
\right]^2
}{
\sum_{t_0,k,g}M_{k}^{\mathrm{grid}}(t_0,g)+\epsilon
}
\label{eq:grid_loss}
\end{equation}
where $M_{k}^{\mathrm{grid}}(t_0,g)$ indicates whether the gridded reference value is valid. The masked loss is evaluated only where the Barnes reference field is valid and constrains the intermediate diagnosis fields and shared regional representation to preserve observed precipitation structure.

A parallel station-rainfall branch predicts the 1 h accumulated precipitation at each supervised station. The corresponding loss is
\begin{equation}
\mathcal{L}_{\mathrm{rain}}
=
\frac{1}{N_{\mathrm{rain}}}
\sum_{t_0,s,k}
M_{s,k}^{\mathrm{rain}}(t_0)
\left[
\widehat{P}_{s,k}^{\mathrm{sta}}(t_0)
-
P_s(t_0+k\,\mathrm{h})
\right]^2
\label{eq:station_rain_loss}
\end{equation}
where $P_s(t_0+k\,\mathrm{h})$ is the observed 1 h accumulated precipitation defined in Section~\ref{subsec:gauge_observations}, $M_{s,k}^{\mathrm{rain}}(t_0)$ indicates the availability of the corresponding observation, and $N_{\mathrm{rain}}$ is the number of valid station--time rainfall observations. This branch retains variations in local rainfall intensity that are not represented by the binary event labels.

The station-related loss and the complete training objective are
\begin{equation}
\begin{aligned}
\mathcal{L}_{\mathrm{station}}
&=
\frac{
2\overline{\mathcal{L}}_{\mathrm{event}}
+
\mathcal{L}_{\mathrm{rain}}
}{3},\\
\mathcal{L}
&=
 w_{\mathrm{grid}}\mathcal{L}_{\mathrm{grid}}
+
 w_{\mathrm{station}}\mathcal{L}_{\mathrm{station}}.
\end{aligned}
\label{eq:total_loss}
\end{equation}
The factor of 2 gives the aggregated event loss twice the weight of station-rainfall regression within the station-related objective. Both $w_{\mathrm{grid}}$ and $w_{\mathrm{station}}$ are set to 1. Together, the three losses supervise complementary aspects of the model: the gridded loss constrains regional precipitation structure, the station-rainfall loss preserves local intensity variation, and the event loss aligns the complete satellite-to-station pathway with the warning target.

\section{Results}
\label{sec:results}

The experiments address two questions: whether the pre-initialization AGRI sequence contains information relevant to subsequent station-neighborhood heavy rainfall, and whether MAGPIE-Net converts this information into local event probabilities more effectively than gridded-output baselines.

\subsection{Experimental Setup and Evaluation Protocol}
\label{subsec:experiment_metrics}

MAGPIE-Net was trained for 15 epochs with a batch size of 2 using the AdamW optimizer. The initial learning rate was $1\times10^{-4}$, and the weight decay was $1\times10^{-3}$. The learning rate was halved after two consecutive epochs without an improvement in validation loss, and the gradient norm was clipped at 5. Model parameters from the epoch with the lowest validation loss in 2022 were used for evaluation on the independent 2023 test set.

All models receive the same four-scan FY-4A AGRI sequence and use the same year-based data split. The adapted Earthformer~\citep{earthformer2022}, PhyDNet~\citep{phydnet2020}, NPM~\citep{park2025npm}, and NowcastNet~\citep{nowcastnet2023} baselines follow the conventional gridded-output pathway shown in Fig.~\ref{fig:paradigm_comparison} and predict gridded 1 h precipitation for the three forecast windows. MAGPIE-Net internally derives the CI features described in Section~\ref{subsec:ci_prior} from the same AGRI sequence and directly predicts station-neighborhood heavy-rainfall event probabilities.

To compare the models at the final warning target, each gridded precipitation forecast is converted into a station-neighborhood event using the same radius-based neighborhood aggregation:
\begin{equation}
\begin{aligned}
\hat{G}_{s,k}^{R,\max}(t_0)
&=
\max_{\substack{g\in\mathcal{G}\\d_{\mathrm{hav}}(g,s)\le R}}
\hat{G}_{k}(t_0,g),\\
\hat{y}_{s,k}^{R,q,\mathrm{grid}}(t_0)
&=
\begin{cases}
1, & \hat{G}_{s,k}^{R,\max}(t_0)\ge 20,\quad q=20,\\
1, & \hat{G}_{s,k}^{R,\max}(t_0)>50,\quad q=50,\\
0, & \mathrm{otherwise}.
\end{cases}
\end{aligned}
\label{eq:grid_to_station_radius_mapping}
\end{equation}
Here, $d_{\mathrm{hav}}(g,s)$ is the haversine distance between grid cell $g$ and target station $s$. The six event definitions combine $R\in\{40,20,10\}\,\mathrm{km}$ with $q\in\{20,50\}\,\mathrm{mm\,h^{-1}}$. The gridded-output baselines retain physical rainfall units, so their neighborhood maxima are compared directly with $q$. The maximum is taken over the full verification radius rather than the smaller local neighborhood typically used for operational grid-to-station mapping. This gives the gridded baselines the same spatial tolerance as the event definition: threshold-exceeding precipitation predicted anywhere within $R$ produces a positive forecast for the target station. MAGPIE-Net directly predicts event probabilities and therefore requires probability decision thresholds. For each event head, the threshold was selected by maximizing CSI on the 2022 validation set, shared across the three forecast windows, and then fixed for evaluation on the independent 2023 test set.

Performance at individual station--time instances is evaluated using the critical success index (CSI), probability of detection (POD), and false alarm ratio (FAR):
\begin{equation}
\mathrm{CSI}
=
\frac{TP}{TP+FP+FN},
\qquad
\mathrm{POD}
=
\frac{TP}{TP+FN},
\qquad
\mathrm{FAR}
=
\frac{FP}{TP+FP}
\label{eq:metrics_csi_pod_far}
\end{equation}
where $TP$, $FP$, and $FN$ denote hits, false alarms, and misses, respectively. The counts are accumulated over all valid station--time instances for each event definition and forecast window.

Episode-level verification uses the primary $40\,\mathrm{km}/20\,\mathrm{mm\,h^{-1}}$ event definition, with each positive label representing an event in the 1 h window ending at its valid time. Because the labels are updated every 15 min, consecutive positives at the same station during sustained rainfall are grouped into one heavy-rainfall episode. The reference time $t_e$ is the end of its first positive window. An episode is considered detected if, no later than $t_e$, the model issues a positive forecast for any 1 h event window belonging to that episode. For each detected episode, the chronologically earliest episode window with a positive forecast and the earliest initialization available for that window are retained; warning lead time is measured from that initialization to $t_e$. Undetected episodes are assigned a lead time of 0 min when calculating the mean episode warning lead time.

Antecedent rainfall characterizes the surface-rainfall state immediately before episode onset. For each episode, $P_{\mathrm{ante}}=P_s^{40,\max}(t_e-1\,\mathrm{h})$ is the maximum 1 h accumulated rainfall within 40 km during the preceding non-overlapping hour. The three below-threshold groups, $P_{\mathrm{ante}}<1$ mm, $1\le P_{\mathrm{ante}}<5$ mm, and $5\le P_{\mathrm{ante}}<20$ mm, contain 9,843, 10,838, and 23,121 episodes, respectively. Another 8,480 episodes have $P_{\mathrm{ante}}\ge20$ mm; in these cases, rainfall had already reached the primary event threshold before at least one subsequent 15-min label fell below the threshold and was followed by renewed exceedance. These episodes are excluded from the stratified analysis because they do not begin from a below-threshold antecedent state.

\subsection{Primary Station-Neighborhood Warning Performance}
\label{subsec:main_performance}

\begin{figure}[htbp]
\centering
\includegraphics[width=\linewidth]{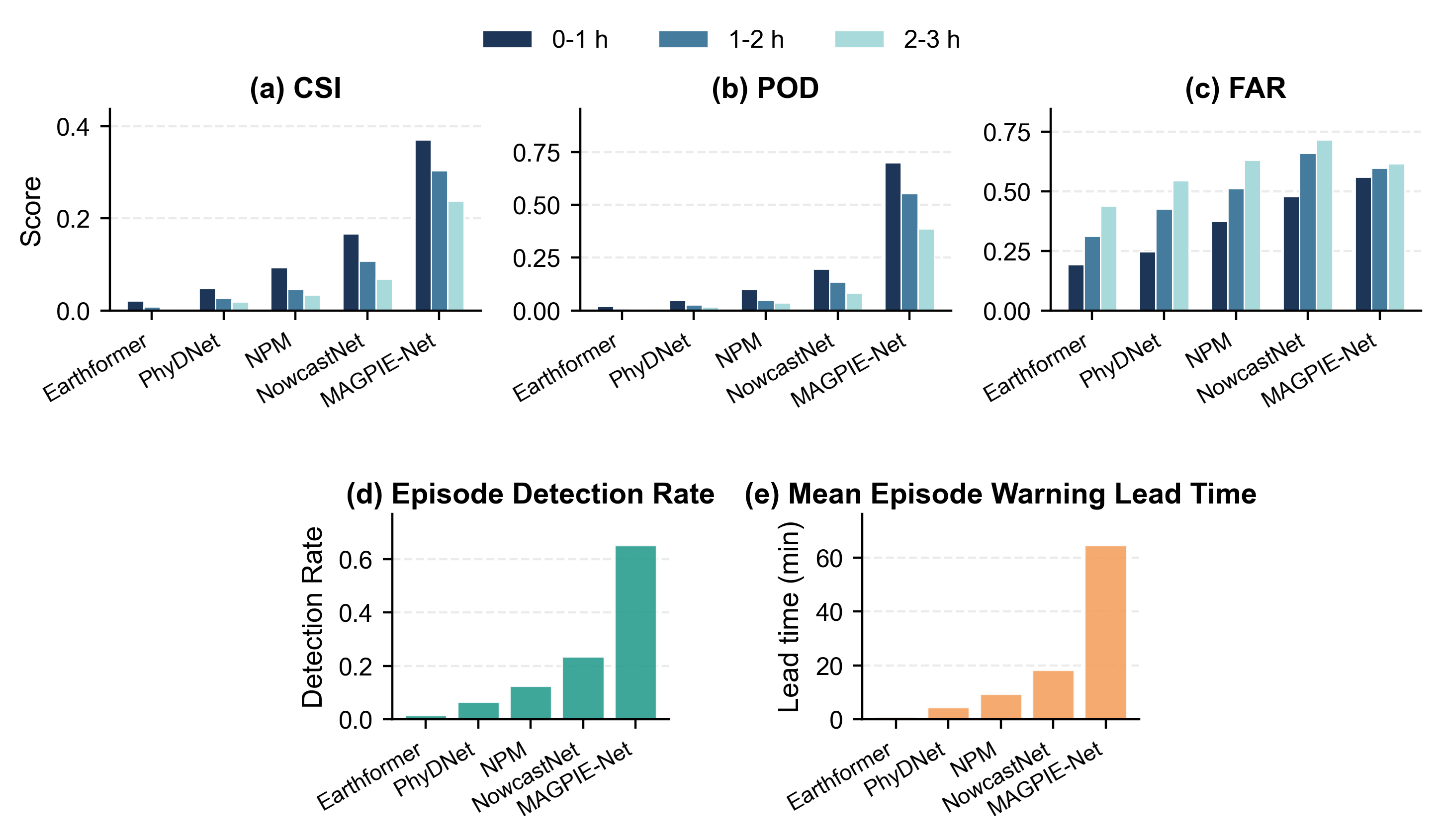}
\caption{Station-neighborhood warning performance under the primary $40\,\mathrm{km}/20\,\mathrm{mm\,h^{-1}}$ event definition. Gridded-output predictions are converted into station-neighborhood events using the same radius-based neighborhood aggregation. (a)--(c) Sample-level CSI, POD, and FAR for the three forecast windows. (d) Episode detection rate. (e) Mean episode warning lead time, with undetected episodes assigned 0 min.}
\label{fig:main_40km20mm}
\end{figure}

As shown in Fig.~\ref{fig:main_40km20mm}, MAGPIE-Net achieves CSI values of \MagpiePrimaryCSIZeroOne{}, \MagpiePrimaryCSIOneTwo{}, and \MagpiePrimaryCSITwoThree{} for the 0--1, 1--2, and 2--3 h forecast windows, respectively, under the primary event definition. The corresponding values for the best-performing gridded-output baseline, NowcastNet, are 0.167, 0.107, and 0.069. All models lose skill as the forecast lead time increases, consistent with growing uncertainty in storm movement, convective growth and decay, and new-cell development. MAGPIE-Net retains a larger proportion of its 0--1 h CSI than NowcastNet, and its relative CSI advantage therefore widens across the three forecast windows. Because all models receive the same four direct AGRI scans, this comparison evaluates how their respective prediction pathways convert cloud evolution into final warning decisions.

The POD and FAR show how this difference changes with forecast lead time. During the first forecast hour, the higher POD of MAGPIE-Net is accompanied by a moderate increase in FAR. Rapid cloud-top cooling and cloud expansion often accompany active convective growth, although some developing clouds later weaken or produce only limited surface rainfall. During the 1--2 h and 2--3 h windows, MAGPIE-Net has both substantially higher POD and lower FAR than NowcastNet. The event-oriented pathway therefore maintains a more favorable balance between missed events and false alarms as uncertainty in rainfall location and intensity increases.

At the episode level, MAGPIE-Net detects \MagpieEpisodeDetectionRate{}\% of the observed heavy-rainfall episodes, compared with \NowcastNetEpisodeDetectionRate{}\% for NowcastNet. The mean episode warning lead time is \MagpieEpisodeMeanLeadTime{} min for MAGPIE-Net and \NowcastNetEpisodeMeanLeadTime{} min for NowcastNet. Together, the higher episode detection rate and longer episode warning lead time show that MAGPIE-Net can use the input AGRI sequence to anticipate impending heavy-rainfall episodes within target-station neighborhoods.

\begin{figure}[htbp]
\centering
\includegraphics[width=\linewidth]{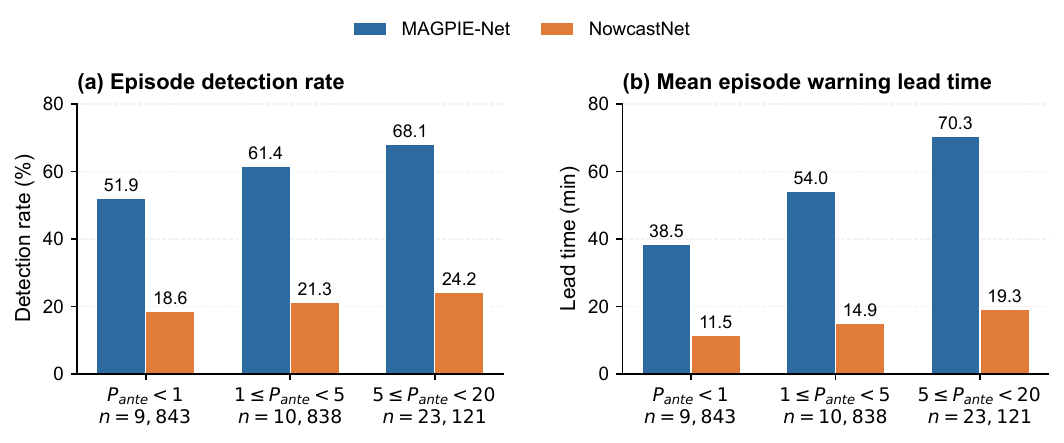}
\caption{Episode-level warning performance stratified by antecedent rainfall within the 40-km neighborhood under the primary event definition. $P_{\mathrm{ante}}$ denotes the maximum 1 h accumulated rainfall during the non-overlapping hour immediately preceding the first threshold-exceeding window. (a) Episode detection rate. (b) Mean episode warning lead time, with undetected episodes assigned 0 min. Numbers beneath the categories indicate the observed episode counts.}
\label{fig:antecedent_rainfall}
\end{figure}

The antecedent-rainfall analysis examines how episode-level performance changes with the amount of surface rainfall present before the first threshold exceedance (Fig.~\ref{fig:antecedent_rainfall}). Both models improve as $P_{\mathrm{ante}}$ increases. This pattern is consistent with greater persistence and spatial organization of the cloud--rain system after surface rainfall begins to develop, providing stronger cues for identifying the subsequent threshold exceedance.

The $P_{\mathrm{ante}}<1$ mm class represents an important early-warning stage in which substantial surface rainfall has not yet developed. Among the \LowAntecedentEpisodeCount{} episodes in this class, MAGPIE-Net detects \MagpieLowAntecedentDetectionRate{}\% and provides a mean episode warning lead time of \MagpieLowAntecedentMeanLeadTime{} min. NowcastNet detects \NowcastNetLowAntecedentDetectionRate{}\% of the same episodes, with a mean episode warning lead time of \NowcastNetLowAntecedentMeanLeadTime{} min. This result shows that MAGPIE-Net can use cloud and moisture evolution observed by AGRI before forecast initialization to anticipate the rapid transition from little or no rainfall to heavy rainfall within the target-station neighborhood.

Across all three antecedent-rainfall classes, the episode detection rate of MAGPIE-Net is approximately 2.8 times that of NowcastNet. Its mean episode warning lead-time advantage increases from 27.0 min for $P_{\mathrm{ante}}<1$ mm to 51.0 min for $5\le P_{\mathrm{ante}}<20$ mm. The advantage is therefore evident across all three below-threshold antecedent states, from near-zero surface rainfall to rainfall approaching the event threshold.

\subsection{Contribution of the Event-Oriented Pathway and GA-SetConv}
\label{subsec:ablation_target_decoder}

Three configurations are evaluated with MAGPIE-Net as the reference. MAGPIE-Net is the complete model. The grid-trained backbone uses the same UNet++ multiscale satellite encoder as MAGPIE-Net but is optimized only against gridded precipitation fields and contains no station-event prediction pathway. Its comparison with MAGPIE-Net evaluates the overall change from gridded-output training to the complete event-oriented satellite-to-station pathway. The bilinear event model retains the CI features, multiscale encoder, auxiliary gridded precipitation diagnosis, and multi-head station-event supervision of MAGPIE-Net, while replacing GA-SetConv with bilinear interpolation. Its comparison with MAGPIE-Net therefore isolates the contribution of GA-SetConv.

Table~\ref{tab:ablation_target_decoder} reports CSI averaged over the three forecast windows. Relative to the grid-trained backbone, MAGPIE-Net increases the mean CSI across the six event definitions from \GridBackboneOverallMeanCSI{} to \MagpieOverallMeanCSI{}, with improvements for every radius--threshold combination. This overall difference reflects the combined effect of the CI features, direct station-event supervision, and trainable GA-SetConv mapping within the complete event-oriented pathway. The improvement is particularly pronounced for the $50\,\mathrm{mm\,h^{-1}}$ events, for which the CSI of the grid-trained backbone is close to zero. Extreme hourly rainfall is often concentrated in localized high-intensity rain cores, so a small error in the position or intensity of a thresholded gridded forecast can change the resulting station-neighborhood classification. Direct event supervision and trainable spatial conversion allow the satellite representation to be optimized against this final classification target.

Replacing bilinear interpolation with GA-SetConv further increases the overall mean CSI from \BilinearEventOverallMeanCSI{} to \MagpieOverallMeanCSI{}, corresponding to a relative gain of \GASetConvRelativeGain{}\%. Improvements occur under all six event definitions and become larger for smaller neighborhoods and the higher rainfall threshold. These stricter definitions are more sensitive to spatial displacement between cloud and rainfall structures and to local geographic conditions. Unlike the fixed local mapping used by bilinear interpolation, GA-SetConv adjusts its axis-aligned east--west and north--south spatial support for each target station and forecast window.

\begin{table}[htbp]
\centering
\caption{Controlled comparison of the event-oriented pathway and grid-to-station decoding. Values are CSI averaged over the three forecast windows.}
\label{tab:ablation_target_decoder}
\footnotesize
\setlength{\tabcolsep}{4pt}
\begin{tabular}{lccccccc}
\toprule
Model &
$40/20$ &
$20/20$ &
$10/20$ &
$40/50$ &
$20/50$ &
$10/50$ &
Mean \\
\midrule
Grid-trained backbone &
0.0456 & 0.0291 & 0.0165 & 0.0007 & 0.0004 & 0.0002 & \GridBackboneOverallMeanCSI{} \\
Bilinear event model &
0.2873 & 0.1802 & 0.1086 & 0.1226 & 0.0703 & 0.0294 & \BilinearEventOverallMeanCSI{} \\
MAGPIE-Net &
\textbf{0.3043} & \textbf{0.1909} & \textbf{0.1183} & \textbf{0.1315} & \textbf{0.0772} & \textbf{0.0376} & \textbf{\MagpieOverallMeanCSI{}} \\
\bottomrule
\end{tabular}

\vspace{1mm}
\begin{minipage}{0.96\linewidth}
\raggedright
\footnotesize The notation $40/20$ denotes $40\,\mathrm{km}/20\,\mathrm{mm\,h^{-1}}$, and the remaining columns follow the same convention.
\end{minipage}
\end{table}

Taken together, the controlled configurations show that the complete event-oriented pathway establishes the main skill gain over gridded-output training, while GA-SetConv provides a consistent additional improvement across all six event definitions.

\subsection{Performance Across Event Definitions and Spatiotemporal Subsets}
\label{subsec:multiscale_extreme_robustness}

\subsubsection{Neighborhood Radius and Rainfall Threshold}
\label{subsubsec:radius_threshold}

\begin{figure}[htbp]
\centering
\includegraphics[width=\linewidth]{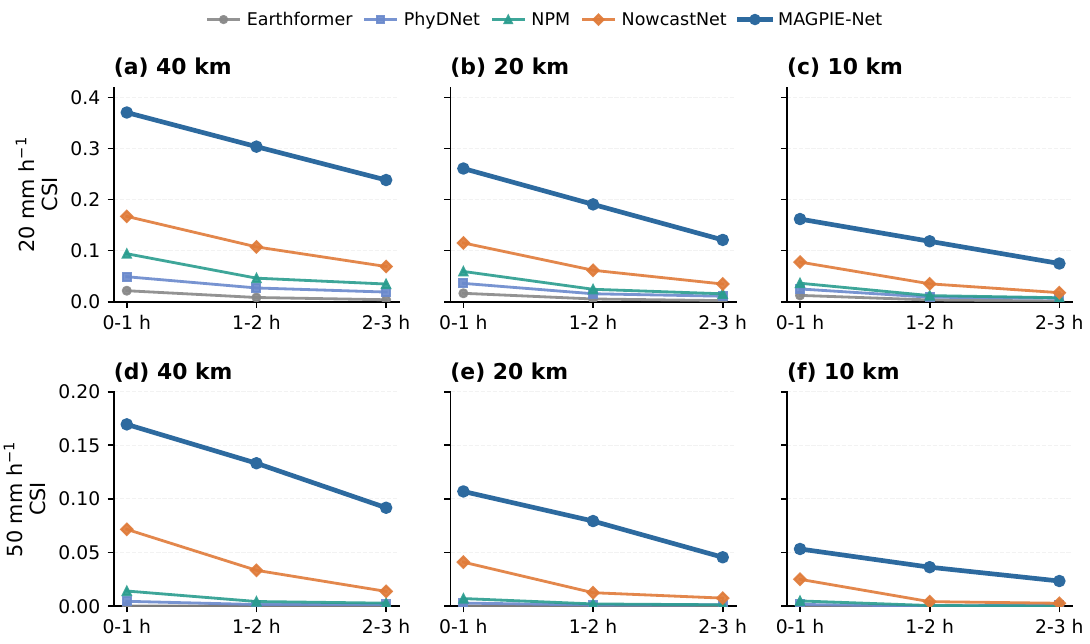}
\caption{Cross-model CSI across neighborhood radii and rainfall thresholds. Gridded-output predictions are evaluated using the same radius-based neighborhood aggregation as the station-neighborhood event definition. Panels (a)--(c) correspond to $20\,\mathrm{mm\,h^{-1}}$, and panels (d)--(f) correspond to $50\,\mathrm{mm\,h^{-1}}$. Columns represent neighborhood radii of 40, 20, and 10 km. Each panel shows the 0--1, 1--2, and 2--3 h forecast windows.}
\label{fig:multiscale_extreme_csi_matrix}
\end{figure}

The results in Fig.~\ref{fig:multiscale_extreme_csi_matrix} indicate that MAGPIE-Net achieves the highest CSI across all 18 combinations of $R=40$, $20$, and $10\,\mathrm{km}$, $q=20$ and $50\,\mathrm{mm\,h^{-1}}$, and the three forecast windows. Its advantage extends from the $40\,\mathrm{km}/20\,\mathrm{mm\,h^{-1}}$ definition to the localized $10\,\mathrm{km}/50\,\mathrm{mm\,h^{-1}}$ definition, showing that the event-oriented pathway maintains a consistent advantage as the warning target requires more precise spatial localization and greater rainfall intensity.

CSI decreases for all models as the neighborhood radius becomes smaller and the rainfall threshold becomes higher, reflecting the increasing difficulty of the task. A 40-km neighborhood accommodates some displacement among the cloud-top structures observed by the satellite, the evolving surface-rainfall area, and the target station. At a radius of 10 km, even small location errors can change the event classification of short-lived convective rain cores. Extreme hourly rainfall at $50\,\mathrm{mm\,h^{-1}}$ is usually confined to a small area and depends on the local co-occurrence of strong ascent, abundant moisture, and efficient cloud microphysical processes. The coldest or most rapidly expanding cloud region does not necessarily coincide with the maximum surface rainfall~\citep{chen2025coldcloud}. The CSI variation under stricter event definitions therefore reflects both event rarity and spatial-scale differences between cloud-top radiative signals and surface rainfall.

\subsubsection{Temporal and Regional Stratification}
\label{subsec:meteorological_stratification}

\begin{figure}[p]
\centering
\includegraphics[width=\textwidth,height=0.84\textheight,keepaspectratio]{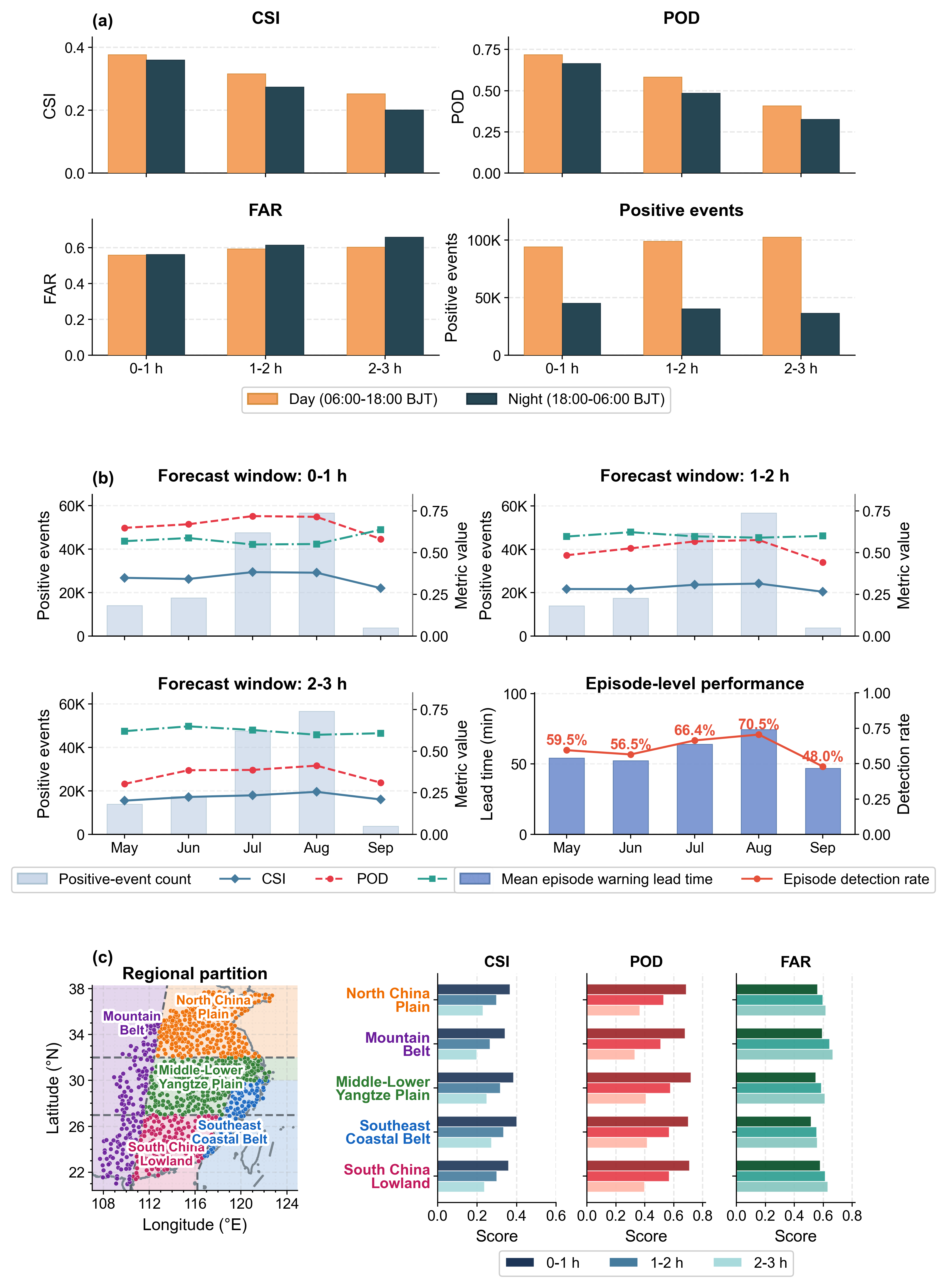}
\caption{Performance of MAGPIE-Net across temporal and regional subsets under the primary event definition. (a) Sample-level CSI, POD, FAR, and positive-event counts for daytime and nighttime samples. (b) Monthly sample-level metrics and positive-event counts for the three forecast windows, together with episode detection rate and mean episode warning lead time. (c) Regional partition and sample-level CSI, POD, and FAR across the five regions.}
\label{fig:meteorological_stratification}
\end{figure}

As shown in Fig.~\ref{fig:meteorological_stratification}, daytime samples, defined as 06:00--18:00 Beijing Time (BJT), have slightly higher CSI and POD than nighttime samples and contain approximately twice as many positive events under the primary event definition. The higher daytime scores coincide with both the larger number of positive events and meteorological conditions that often favor locally initiated convection~\citep{chen2013sdhr,luo2017scmrex}. Successive infrared observations can track the rapidly evolving cloud structures associated with such development.

After sunset, MAGPIE-Net retains station-neighborhood warning performance across all three forecast windows. Infrared and water-vapor channels continue to observe cloud-top and moisture evolution in nocturnal mesoscale convective systems, coastal convection, and tropical-cyclone rainbands~\citep{chen2013sdhr,lonfat2004tc}. The observed day--night differences occur alongside changes in event frequency, convective organization, and the relationship between upper-cloud evolution and surface rainfall.

Performance is strongest in July and August, when positive events are most frequent and the episode detection rates reach 66.4\% and 70.5\%, respectively. These months combine larger positive-event samples with frequent deep convection, organized mesoscale systems, and tropical-weather influences that often produce persistent and spatially coherent upper-cloud structures~\citep{chen2013sdhr,lonfat2004tc}. May, June, and September contain fewer positive events and a broader mixture of frontal, warm-sector, and terrain-influenced rainfall~\citep{luo2017scmrex,houze2012orographic}. Across all five months, episode detection ranges from 48.0\% in September to 70.5\% in August, and the mean warning lead time ranges from 47.0 to 74.5 min.

Regional differences are moderate. The Southeast Coastal Belt has the highest CSI and lowest FAR across all three forecast windows, while its POD remains comparable to the leading regional values. The Mountain Belt has lower POD and higher FAR at longer lead times. These contrasts are consistent with regional differences among organized coastal and tropical-cyclone rainbands, terrain-influenced precipitation, and warm-sector rainfall, which can alter the spatial correspondence between upper-cloud evolution and surface rainfall~\citep{luo2017scmrex,lonfat2004tc,houze2012orographic}. Despite these regional variations, the decline in warning skill with forecast lead time is similar across all five partitions.

\subsection{CI-Feature Ablation and Spatial Attribution}
\label{subsec:ci_masking_attribution}

\begin{figure}[htbp]
\centering
\includegraphics[width=\linewidth]{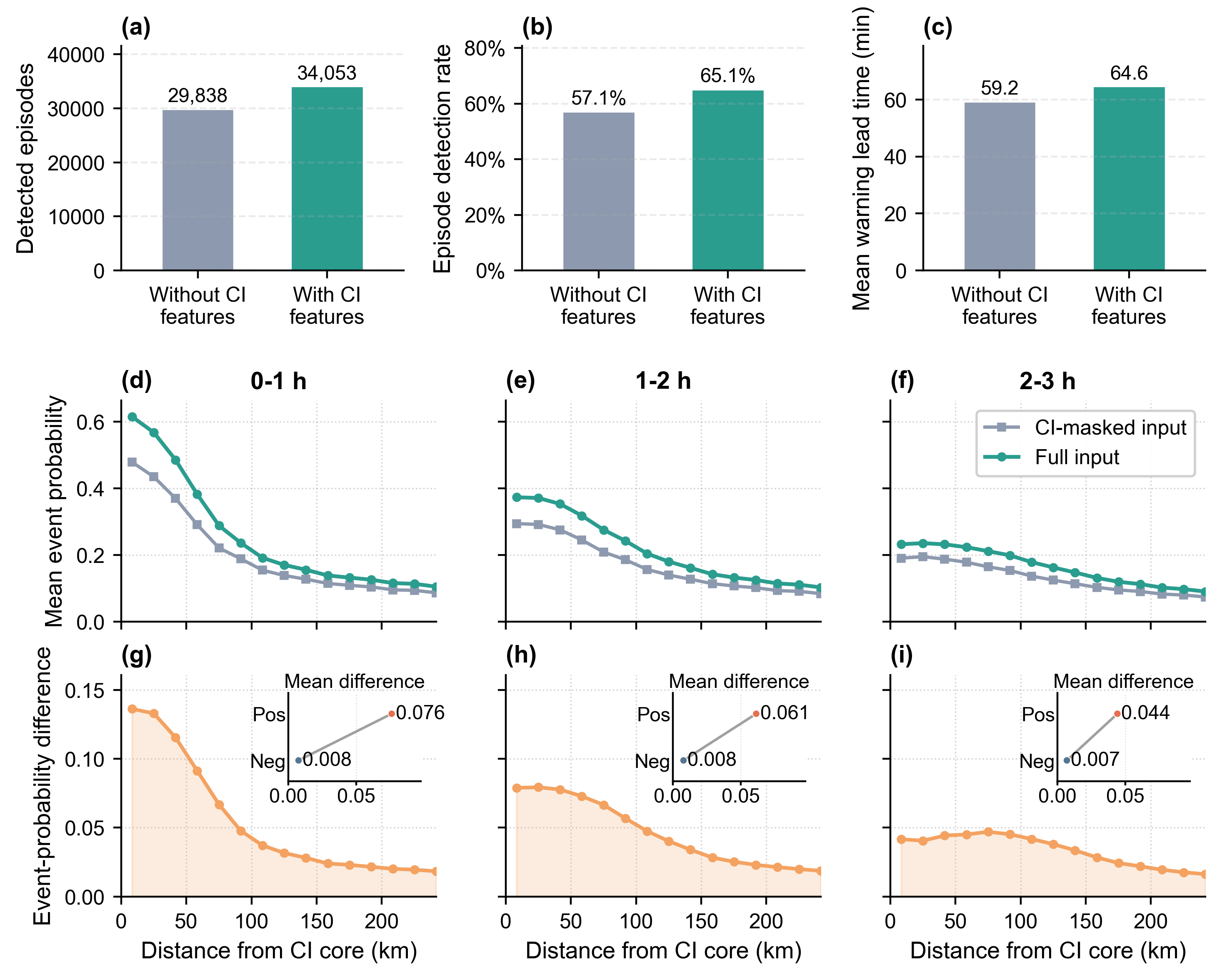}
\caption{CI-feature ablation and inference-time spatial attribution under the primary event definition. (a)--(c) Detected episode count, episode detection rate, and mean warning lead time for independently trained MAGPIE-Net models without and with CI features. (d)--(f) CI-centered station-neighborhood event-probability profiles obtained with the full input and the CI-masked input. (g)--(i) Event-probability differences between the full and CI-masked inputs; the insets compare the mean differences for positive and negative station--time events.}
\label{fig:ci_combined}
\end{figure}

Two experiments separated the contribution of CI features to warning performance from their spatial influence on the trained model. In the retraining ablation, MAGPIE-Net was trained without the five CI feature channels while retaining the same input AGRI sequence, event heads, and GA-SetConv decoder. The model with CI features detected \MagpieDetectedEpisodeCount{} episodes, with an episode detection rate of \MagpieEpisodeDetectionRate{}\% and a mean warning lead time of \MagpieEpisodeMeanLeadTime{} min. The retrained model without CI detected \NoCIDetectedEpisodeCount{} episodes, with a detection rate of \NoCIEpisodeDetectionRate{}\% and a mean warning lead time of \NoCIEpisodeMeanLeadTime{} min (Fig.~\ref{fig:ci_combined}a--c). Thus, including CI features allowed the model to detect additional heavy-rainfall episodes and provide earlier warnings.

An inference-time masking experiment then examined where CI information affected the output of the trained model. MAGPIE-Net was evaluated with the full input and the CI-masked input while its parameters and all other inputs remained unchanged. The event-probability difference was largest near the CI core and decreased with distance (Fig.~\ref{fig:ci_combined}d--i). This response was strongest in the 0--1 h window and weakened with forecast lead time. The mean difference for positive station--time events was approximately six to ten times that for negative events. The localized radial response and the stronger response for positive events indicate that the model used CI features selectively rather than raising probabilities uniformly around every CI candidate. This pattern is consistent with the permissive CI construction, in which the candidate masks identify possible regions of rapid cloud development and the surrounding AGRI context helps distinguish candidates associated with subsequent heavy rainfall.

\subsection{Representative Heavy-Rainfall Cases}
\label{subsec:case_study}

\subsubsection{Rapid Local Convective Development From Near-Zero Antecedent Rainfall}
\label{subsubsec:convective_onset_case}

\begin{figure}[htbp]
\centering
\includegraphics[width=\linewidth]{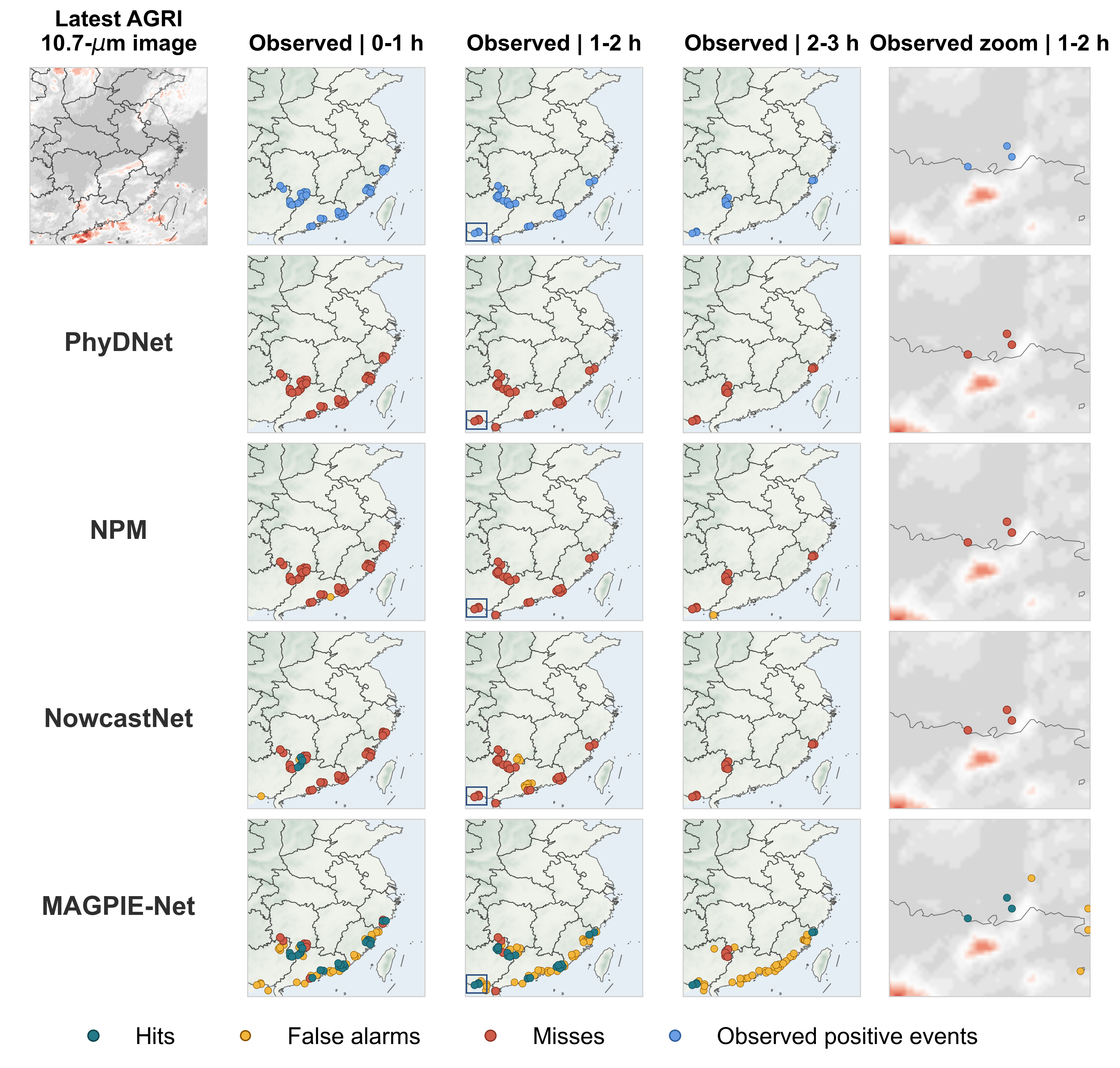}
\caption{Station-neighborhood event predictions for a localized convective heavy-rainfall episode over coastal southern Guangxi, initialized at 22:00 UTC on 10 August 2023 (06:00 BJT on 11 August). The top row shows the latest input AGRI $10.7\,\mu\mathrm{m}$ image at 21:45 UTC (05:45 BJT), observed positive events in the three forecast windows, and an enlarged view of the newly positive component during 1--2 h. The rightmost column uses the same AGRI $10.7\,\mu\mathrm{m}$ background for all rows. The remaining rows show predictions from PhyDNet, NPM, NowcastNet, and MAGPIE-Net. EarthFormer, omitted for clarity, also missed all three stations in the highlighted component. Cyan, orange, red, and blue markers denote hits, false alarms, misses, and observed positive events, respectively.}
\label{fig:case_study_convective_onset}
\end{figure}

Figure~\ref{fig:case_study_convective_onset} presents a localized convective heavy-rainfall episode over coastal southern Guangxi, initialized at 22:00 UTC on 10 August 2023 (06:00 BJT on 11 August). During the non-overlapping hour before the first threshold-exceeding window, the maximum 1 h accumulated rainfall across the 40-km neighborhoods of the three stations in the highlighted component was only $0.5\,\mathrm{mm}$. The latest input AGRI image showed a localized cold-cloud feature near the Beibu Gulf coast. All three stations in the highlighted component then became positive during the 1--2 h forecast window. The case therefore represents a rapid transition from near-zero antecedent surface rainfall to localized station-neighborhood heavy rainfall.

MAGPIE-Net detected all three newly positive stations, whereas PhyDNet, NPM, EarthFormer, and NowcastNet missed all three. Within the enlarged local domain, MAGPIE-Net produced three hits, no misses, and four false alarms, corresponding to a local CSI of 0.429; none of the gridded-output baselines produced a hit. The case provides a direct example of MAGPIE-Net identifying localized heavy-rainfall development from the evolving cloud field before substantial surface rainfall had formed.

\subsubsection{Organized Tropical-Cyclone Rainfall During Typhoon Doksuri}
\label{subsubsec:doksuri_case}

\begin{figure}[htbp]
\centering
\includegraphics[width=\linewidth]{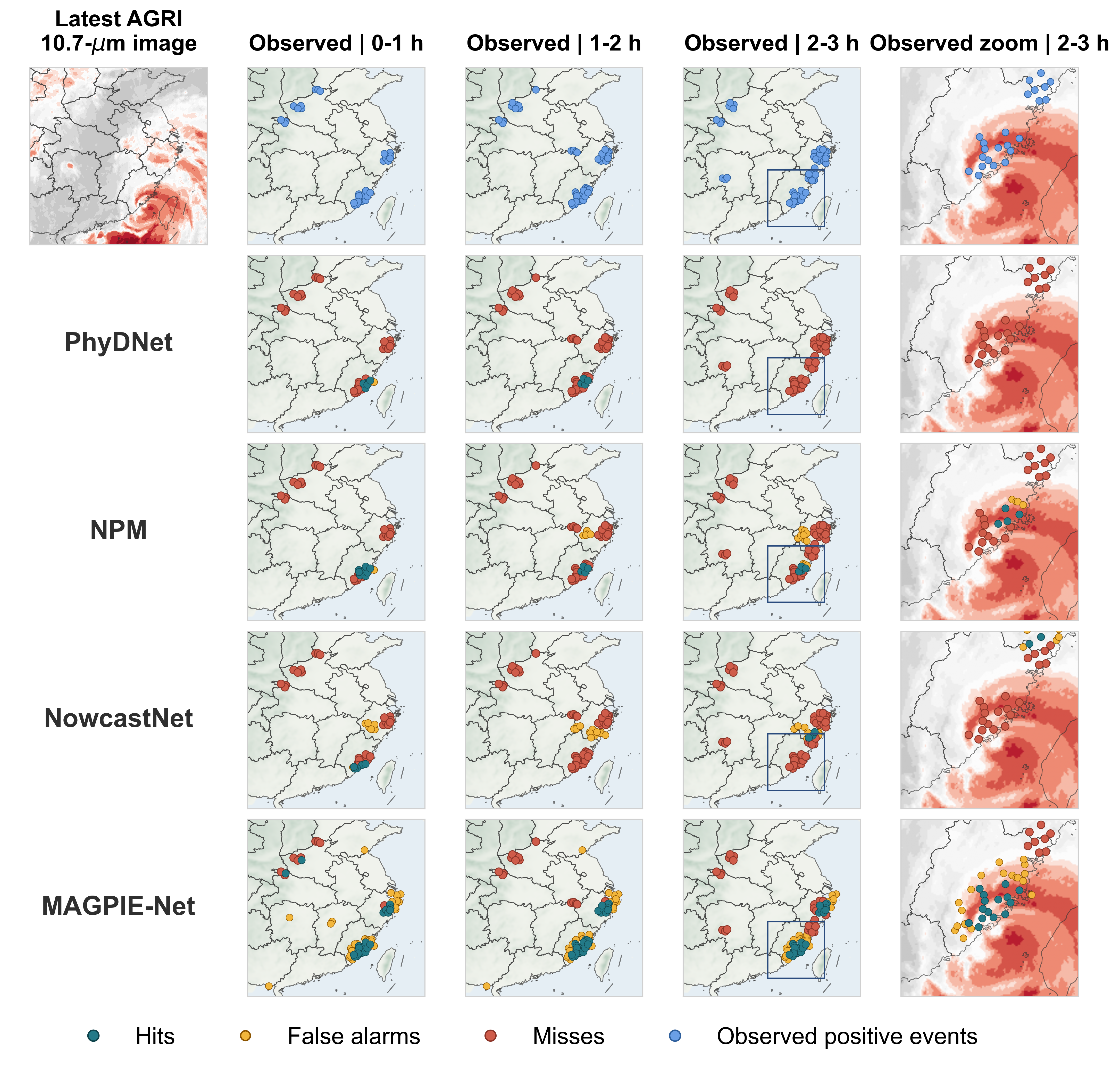}
\caption{Station-neighborhood event predictions during the landfall of Typhoon Doksuri, initialized at 00:00 UTC (08:00 BJT) on 28 July 2023. The top row shows the latest input AGRI $10.7\,\mu\mathrm{m}$ image, observed positive events in the three forecast windows, and an enlarged coastal view during 2--3 h. The rightmost column uses the same AGRI $10.7\,\mu\mathrm{m}$ background for all rows. The remaining rows show predictions from PhyDNet, NPM, NowcastNet, and MAGPIE-Net. Cyan, orange, red, and blue markers denote hits, false alarms, misses, and observed positive events, respectively.}
\label{fig:case_study}
\end{figure}

Figure~\ref{fig:case_study} presents a contrasting, spatially organized rainfall case during the landfall of Typhoon Doksuri on 28 July 2023. The forecast was initialized at 00:00 UTC (08:00 BJT), approximately two hours before landfall near Jinjiang, Fujian. At initialization, the AGRI image showed a mature central cloud shield and spiral cloud bands approaching southeastern China. Over the following three hours, the observed station-neighborhood events expanded along the Fujian--Zhejiang coast as the tropical-cyclone circulation and associated rainbands moved inland.

The gridded-output models produced fragmented warnings and missed substantial parts of the coastal event band. MAGPIE-Net produced a more continuous band of hits and achieved the highest case-level CSI in all three forecast windows. The difference was most pronounced during 1--2 h and 2--3 h, as the affected area expanded along the coast. Some false alarms occurred beneath the broad cloud shield but outside the strongest surface rainbands. Their spatial distribution is consistent with the asymmetric rainfall structure of tropical cyclones~\citep{lonfat2004tc} and the displacement between convective cold-cloud features and surface-rainfall maxima~\citep{chen2025coldcloud}. Across the three windows, MAGPIE-Net followed the observed coastal expansion of the rainband more continuously than the gridded-output models.

The two cases span complementary stages of cloud--rain evolution. The southern Guangxi case follows localized heavy-rainfall development from near-zero antecedent surface rainfall, whereas the Doksuri case follows the inland expansion of a mature and organized tropical-cyclone rainband. MAGPIE-Net identified the newly positive southern Guangxi component missed by all gridded-output baselines and more continuously captured the coastal expansion of the Doksuri rainband.

\clearpage
\section{Conclusions}
\label{sec:conclusion}

We developed MAGPIE-Net, an event-oriented satellite-to-station model that uses multitemporal FY-4A AGRI observations to predict station-neighborhood heavy-rainfall events at lead times of 0--3 h. The model combines convection-initiation feature construction, multiscale satellite encoding, an auxiliary gridded precipitation diagnosis, and geographically adaptive grid-to-station decoding. This design preserves the regional organization of cloud- and rainfall-related features, while the station-neighborhood event target directly guides their conversion into local station-neighborhood event probabilities.

In an independent evaluation over central and eastern China during the 2023 warm season, MAGPIE-Net outperformed the gridded-output baselines across all three forecast windows under the primary event definition. This advantage persisted under smaller neighborhood radii and the $50\,\mathrm{mm\,h^{-1}}$ threshold. When consecutive events at the same station were grouped into heavy-rainfall episodes, MAGPIE-Net achieved an episode detection rate of \MagpieEpisodeDetectionRate{}\% and a mean episode warning lead time of \MagpieEpisodeMeanLeadTime{} min, compared with \NowcastNetEpisodeDetectionRate{}\% and \NowcastNetEpisodeMeanLeadTime{} min for the best-performing gridded-output baseline. During the critical early-warning stage, when rainfall during the preceding hour remained below 1 mm within the 40-km neighborhood, MAGPIE-Net still detected \MagpieLowAntecedentDetectionRate{}\% of the subsequent episodes and provided a mean episode warning lead time of \MagpieLowAntecedentMeanLeadTime{} min. Together, these results show that continuous AGRI observations of cloud and moisture evolution can provide warning-relevant information for station-neighborhood heavy rainfall before substantial surface rainfall develops.

The configuration experiments clarify how MAGPIE-Net converts satellite information into station-neighborhood warnings. The complete event-oriented configuration produced the largest improvement over the gridded-output pathway, while GA-SetConv provided an additional gain through station- and lead-dependent spatial aggregation. Relative to the retrained no-CI model, including CI features increased the episode detection rate and mean warning lead time. Inference-time masking further showed that the probability response to CI information was concentrated near rapidly developing clouds.

Performance is lowest for spatially localized extreme-rainfall events under the smaller-neighborhood and higher-intensity definitions. AGRI cloud-top and water-vapor observations do not directly resolve low-level moisture convergence, warm-rain microphysics, or terrain-induced near-surface processes, and cloud-top features can be displaced from surface rainfall. Future work can combine AGRI observations with radar echoes, lightning, recent station rainfall, and atmospheric-state information to complement AGRI with information about the near-surface environment and hydrometeor structure. Evaluation across other climate regions and geostationary satellite sensors can further assess model transferability and probability calibration.

Taken together, these results show that, compared with conventional gridded-output modeling, the event-oriented satellite-to-station pathway more effectively translates multitemporal AGRI observations into local short-duration heavy-rainfall warnings.

\bibliographystyle{plainnat}
\bibliography{references}

\end{document}